%% file: acl_latex.tex
\documentclass[11pt]{article}

\usepackage[preprint]{acl}

\usepackage{times}
\usepackage{latexsym}

\usepackage[T1]{fontenc}

\usepackage[utf8]{inputenc}

\usepackage{microtype}

\usepackage{inconsolata}

\usepackage{graphicx}
\usepackage{float}
\usepackage{capt-of}
\usepackage{amsmath}
\usepackage{amssymb}
\usepackage{amsthm}

\newtheorem{corollary}{Corollary}
\newtheorem{proposition}{Proposition}
\usepackage{booktabs}
\usepackage{array}
\usepackage{colortbl}
\usepackage{tabularx}
\usepackage{titletoc}
\usepackage{enumitem}
\usepackage{pifont}
\definecolor{sage}{HTML}{99BAB9}
\definecolor{lavender}{HTML}{D6CDEA}
\definecolor{cream}{HTML}{FEFBEB}
\colorlet{oursrow}{sage!20}
\definecolor{band}{HTML}{F8F3E7}
\definecolor{bandedge}{HTML}{E2D6B8}
\usepackage[most]{tcolorbox}
\definecolor{boxedge}{HTML}{C9DCDB}
\newtcolorbox{resultbox}[1]{enhanced,breakable,colback=white,colframe=boxedge,coltitle=black,
  fonttitle=\bfseries,title=#1,boxrule=1pt,arc=3pt,left=4pt,right=4pt,top=1pt,bottom=1pt,
  toptitle=1pt,bottomtitle=1pt,before skip=5pt,after skip=5pt,
  before upper={\setlength{\topsep}{0pt}}}
\newtcblisting{promptbox}[1]{enhanced,breakable,listing only,title=#1,colback=black!3,colframe=black!18,
  colbacktitle=black!12,coltitle=black,fonttitle=\bfseries\small,boxrule=0.5pt,arc=2pt,left=3pt,right=3pt,
  top=2pt,bottom=2pt,before skip=6pt,after skip=8pt,
  listing options={basicstyle=\ttfamily\scriptsize,breaklines=true,breakindent=0pt,columns=fullflexible,keepspaces=true,aboveskip=0pt,belowskip=0pt}}
\newcommand{\hlnode}[2]{\tikz[remember picture,baseline=(#1.base)]\node[inner sep=0pt,outer sep=0pt](#1){#2};}
\newcommand{\hlcolumn}[3][3.5pt]{\tikz[remember picture,overlay]\draw[bandedge,line width=0.4pt,fill=band,rounded corners=3pt,blend mode=multiply]
  ($(#2T.north)+(-#3,#1)$) rectangle ($(#2B.south)+(#3,-3pt)$);}
\usepackage{tikz}
\usetikzlibrary{calc}
\definecolor{hoplight}{HTML}{EBE6F6}
\definecolor{finallight}{HTML}{C9DCDB}
\definecolor{nodeedge}{HTML}{5A5A5A}
\definecolor{factgray}{HTML}{767676}
\newcommand{\enode}[2]{\tikz[baseline={([yshift=-0.55ex]n.center)}]\node[circle,draw=nodeedge,line width=0.5pt,fill=#2,inner sep=0.6pt,minimum size=1.15em,font=\footnotesize](n){$e_{#1}$};}
\definecolor{histlight}{HTML}{F2F2F2}
\definecolor{nonimglight}{HTML}{F5E6E9}
\definecolor{withheld}{HTML}{A9A6A0}
\definecolor{histstrong}{HTML}{D3D2CF}
\definecolor{examstrong}{HTML}{ADD5B4}
\definecolor{nonimgstrong}{HTML}{EEBBA5}
\definecolor{imgstrong}{HTML}{9CBCE4}

\newcommand{\cnum}[2]{\tikz[baseline={([yshift=-0.55ex]n.center)}]\node[rectangle,rounded corners=1pt,draw=nodeedge,line width=0.4pt,fill=#2,inner sep=0.3pt,minimum size=1.05em,font=\scriptsize](n){#1};}
\newcommand{\stateicon}[4]{\tikz[baseline=-0.6ex]{%
  \foreach \v/\c/\i in {#1/histstrong/0,#2/examstrong/1,#3/nonimgstrong/2,#4/imgstrong/3}{%
    \ifnum\v=1 \draw[nodeedge,line width=0.4pt,fill=\c,rounded corners=0.6pt] (\i*0.72em,-0.3em) rectangle ++(0.6em,0.6em);%
    \else \draw[withheld,line width=0.4pt,dash pattern=on 1pt off 0.7pt,rounded corners=0.6pt] (\i*0.72em,-0.3em) rectangle ++(0.6em,0.6em);\fi}}}
\newcommand{\yesmark}{\checkmark}

\newcommand{\nomark}{\leavevmode{\color{factgray}--}}
\newcommand{\enodenone}{\ensuremath{\varnothing}}
\newcommand{\enodesm}[2]{\tikz[baseline={([yshift=-0.55ex]n.center)}]\node[circle,draw=nodeedge,line width=0.4pt,fill=#2,inner sep=0.3pt,minimum size=1.0em,font=\scriptsize](n){$e_{#1}$};}

\newcommand{\dnodesm}[1]{\tikz[baseline={([yshift=-0.55ex]n.center)}]\node[rectangle,rounded corners=1pt,draw=withheld,line width=0.4pt,fill=histlight,inner sep=0.3pt,minimum size=1.0em,font=\scriptsize](n){$d_{#1}$};}
\newcommand{\sd}[1]{\,{\scriptsize$\pm$#1}}

\newcommand{\bestm}[2]{\textbf{#1}\sd{#2}}
\newcommand{\ours}{SuffiCE}
\newcommand{\tok}[1]{{\setlength{\fboxsep}{1.2pt}\colorbox{black!8}{\texttt{#1}}}}

\title{Each Insufficient in Its Own Way: \\ Calibrating Language Models Across Varying Evidence States}

\author{
\textbf{Jiayi Li}$^{\star}$ \quad
\textbf{Dennis L. Shung}$^{\diamond}$ \quad
\textbf{Bradly C. Stadie}$^{\star}$ \\
\\[-0.8em]
$^{\star}$Northwestern University \quad
$^{\diamond}$Mayo Clinic
}

\begin{document}
\maketitle
\begin{abstract}
How can a language model know whether it has enough evidence to answer? Language models tend to answer even when the evidence they have is insufficient. Deciding whether to answer or gather more evidence requires a calibrated estimate of how likely an answer is to be correct with the available evidence. However, most calibration work treats the input as fixed, without asking how this estimate should change with the evidence. We introduce \emph{evidence-state calibration}, which requires a model's confidence before answering to match its probability of answering correctly and to change accordingly whenever the evidence changes this probability. We propose \ours{} (\textbf{Suffi}ciency \textbf{C}onfidence \textbf{E}stimation), which fine-tunes the confidence already present in a model's next-token distribution to match how often the model answers correctly from each evidence state. Across clinical diagnosis and multi-hop question answering with four models from two families, \ours{} reduces the Brier score by up to 60\% and increases AUROC by up to 0.16. Its confidence matches the empirical success rate and responds not only to the amount of evidence but also to which evidence is present, showing that models can learn to judge whether their current evidence is enough to answer.
\end{abstract}

\input{sections/introduction.tex}
\input{sections/related_works.tex}
\input{sections/method.tex}
\input{sections/experimental_setup.tex}
\input{sections/results.tex}

\input{sections/conclusion.tex}

\section*{Limitations}

\ours{} shows strong calibration and evidence sensitivity across four answering models and three datasets, but it has three limitations. First, we preconstruct evidence states rather than gather evidence in live multi-turn interactions, a controlled setting that lets us attribute differences in success and confidence to the evidence itself. Second, to limit cost, we estimate each target from ten sampled answers per state, and more samples would give more precise targets. Third, \ours{} has not been tested in real clinical scenarios, and its confidence should not replace clinical judgment.

\bibliography{custom}

\clearpage
\appendix
\startcontents[appendix]
\printcontents[appendix]{}{1}{\section*{Appendix Contents}\setcounter{tocdepth}{2}}
\clearpage

\input{sections/appendix.tex}

\end{document}

%% file: sections/introduction.tex
\section{Introduction}
\label{sec:introduction}

Before answering a question, one must judge whether the information at hand is enough to answer it. In high-stakes settings, a physician who diagnoses before gathering enough information may miss the real disease. For example, a patient who has been tired for two months may seem iron deficient, but a careful clinician would order more blood tests, which may reveal celiac disease, an autoimmune cause of poor iron absorption. 

\begin{figure}[t]
    \centering
    \includegraphics[width=0.8\columnwidth]{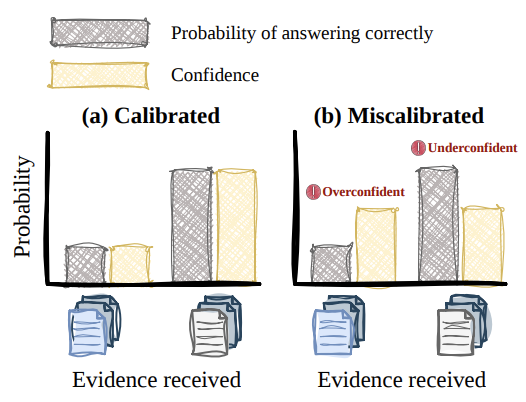}
    \caption{Given a question with two different sets of evidence, the model's probability of answering correctly (gray) differs, and its confidence before answering (yellow) should match it for each. (a)~Calibrated confidence matches both. (b)~Miscalibrated confidence mismatches the dynamic of the true probability.}
    \label{fig:information_construction}
\end{figure}

Language models facing insufficient information, however, often reach confident but incorrect conclusions~\citep{kirichenkoabstentionbench2025,yin-etal-2023-large}. A common remedy for such overconfidence is to abstain below a confidence threshold~\citep{wen-etal-2025-know}. This works only if the model is calibrated, so that answers given x\% confidence are correct about x\% of the time~\citep{guo2017calibration,geng-etal-2024-survey}. To achieve this, existing methods train language models to provide calibrated confidence estimates for generated answers~\citep{kapoor2024LLM,bani2026rewarding}.

However, most calibration work holds the evidence fixed for each question, so it can check whether confidence is calibrated but not whether it varies correctly as a function of the evidence. In interactive settings, the available information evolves as models ask clarifying questions~\citep{zhang-choi-2025-clarify} or seek additional information~\citep{li2024mediq}. At each turn, the model must decide whether to answer or keep gathering evidence.

To make this decision, a model needs a confidence estimated from the available evidence alone, before it produces any answer. We call the evidence available to a model at a given moment its evidence state. We then define \emph{evidence-state calibration} as requiring the confidence before answering to match the probability of answering correctly, so that whenever the evidence changes this probability, the confidence changes by the same amount (Figure~\ref{fig:information_construction}). 

Testing this requires observing the same problem under different evidence states. Rather than contrasting the extremes of complete and missing information~\citep{kirichenkoabstentionbench2025} or following a single fixed sequence of accumulating evidence~\citep{zhang-etal-2026-confidence-estimation}, we construct granular states for each problem, spanning minimal to complete evidence and including different combinations of information. Granularity matters because sufficient evidence states are alike, but each insufficient one is insufficient in its own way.\footnote{After the opening line of Leo Tolstoy's \emph{Anna Karenina} (1878): ``Happy families are all alike; every unhappy family is unhappy in its own way.''} Some insufficient states can still be answered from what the model already knows~\citep{jorenSufficientContext2025}, while others lack the one piece of evidence the answer depends on. We therefore measure each state's sufficiency by how often the model answers correctly from it, not by whether the required facts are present, and teach the model to anticipate this success rate before answering.

With these success rates as targets, we propose \ours{} (\textbf{Suffi}ciency \textbf{C}onfidence \textbf{E}stimation), to our knowledge the first method that trains a model's confidence before answering to be evidence-state calibrated. It reads the confidence directly from the model's own next-token distribution. The confidence is the model's probability of choosing to answer instead of asking for more information. No additional classifier or generated number is needed. This avoids an ambiguity in verbalized confidence. When a model states that it is 80\% confident, it is unclear whether to trust the stated number or the probability the model assigned to generating it. We instead train the model's own probability directly to match the success rate across our diverse evidence states. The result is a model that can tell, before answering, whether what it currently knows is enough to answer correctly.

We evaluate on two clinical diagnosis datasets, MIMIC~\citep{hager2024evaluation} and MedQA~\citep{jin2021disease}, and the multi-hop question answering dataset MuSiQue~\citep{trivedi-etal-2022-musique}. Across three Qwen models from 4B to 27B parameters and Kimi K2.6, \ours{} improves calibration for every model and dataset, reducing the Brier score of the base logit by up to 60\% and increasing its AUROC by up to 0.16. Its confidence also follows the evidence state. An estimator trained on complete evidence alone falls short on every metric, so these gains come from training across evidence states. In practice, answering only when \ours{} is confident gives fewer wrong answers than any baseline.

%% file: sections/related_works.tex
\section{Related Work}
\label{sec:related_works}

Deciding whether to answer or gather more evidence requires a calibrated judgment of whether the current evidence suffices. This connects three lines of work: interactive information seeking explores how models acquire evidence, information sufficiency and unanswerability examine whether an input supports an answer, and confidence calibration studies whether confidence reflects actual accuracy.

\paragraph{Interactive information seeking.}
In an interactive setting, a model can gather more information before answering by asking clarifying questions~\citep{zhang-choi-2025-clarify,li2024mediq}, retrieving documents~\citep{jiang-etal-2023-active,asai2024selfrag}, or ordering diagnostic tests and searching for medical evidence~\citep{schmidgall2026agentclinic,dang2026KnowGuard}. Deciding when enough has been gathered typically relies on a confidence or uncertainty signal~\citep{jiang-etal-2023-active,zhang-choi-2025-clarify,li2024mediq}, and MediQ shows that improving confidence estimation can enhance interactive diagnostic accuracy~\citep{li2024mediq}. Our approach aims to provide such a signal, estimating before answering how likely the current evidence is to yield a correct answer. Closest to our work, LA-CDM trains multi-turn diagnosis with a verbalized confidence that guides whether to request another test, but this confidence evaluates a hypothesis the model has already generated~\citep{bani2026lacdm}. We instead estimate success before generating an answer and train this estimate across controlled evidence states of the same problem, capturing how changes in evidence affect the likelihood of success.

\paragraph{Information sufficiency and unanswerability.}
Whether an input contains enough information to answer a question has been widely studied through unanswerability~\citep{rajpurkar-etal-2018-know,yin-etal-2023-large}. Benchmarks test whether models abstain when information is insufficient~\citep{kirichenkoabstentionbench2025}, including settings where key evidence is removed from otherwise answerable questions~\citep{trivedi-etal-2022-musique,machcha-etal-2026-knowing}. Detection methods assess sufficiency using model hidden states~\citep{slobodkin-etal-2023-curious,lavi-etal-2026-detecting} or by identifying missing information~\citep{jain-garimella-2026-knowing}, and training methods teach models to abstain when information is insufficient~\citep{zhai-etal-2026-abstain}. A shared limitation is that sufficiency is a binary label attached to the input, so a partially informative input counts as either sufficient or not, regardless of which model answers it. \citet{jorenSufficientContext2025} take a first step, showing that models often answer correctly from insufficient context. Whether evidence is enough depends on the model answering, so we define sufficiency by the model's success, how often it answers correctly from each controlled evidence state.

\paragraph{Confidence calibration.}
A model is calibrated when its confidence matches its accuracy~\citep{guo2017calibration}. Training-free estimates based on verbalized judgments~\citep{tian-etal-2023-just,xiongCanLLMsExpress2024}, sampled-answer consistency~\citep{kuhn2023semantic}, or token probabilities~\citep{kadavath2022language} can be informative, but calibration varies across tasks~\citep{vashurin-etal-2025-benchmarking}. Trained methods estimate answer correctness~\citep{kapoor2024LLM}, verbalize calibrated confidence~\citep{lin2022teaching,bani2026rewarding,damani2026beyond,li2025ConfTuner}, or predict success before answering~\citep{kadavath2022language,lugoloobi2026encode}. However, they do not explicitly train confidence to track success across changing evidence states. \citet{zhang-etal-2026-confidence-estimation} evaluate confidence as hints accumulate in a fixed sequence but do not train on these changing states. We instead train and evaluate confidence across controlled evidence states, varying both the amount and type of evidence to track changes in empirical success (Appendix~\ref{app:related_comparison}).

%% file: sections/method.tex
\section{Method}
\label{sec:method}

Our method, \ours{}, has three stages. We first construct multiple evidence states for each problem (\S\ref{sec:info_states}), then estimate a fixed answering policy's success rate on each state from repeated answers (\S\ref{sec:success_labels}), and finally train an estimator to predict this rate from the state alone, before answering (\S\ref{sec:estimator}). 

\subsection{Problem Formulation}
\label{sec:formulation}

Problem $i$ consists of a question $q_i$, a pool of evidence $E_i$, and a gold answer $y^*_i$. An \emph{evidence state} $x=(q_i,I)$ pairs the question with the evidence $I\subseteq E_i$ available to the model.

A fixed answering policy $\pi_0$ samples an answer $\hat y\sim\pi_0(\cdot\mid x)$, which a grader scores as $g(\hat y,y^*_i)\in\{0,1\}$. The \emph{success probability} of an evidence state is
\begin{equation}
    p^*(x)=\Pr_{\hat y\sim\pi_0(\cdot\mid x)}\bigl[g(\hat y,y^*_i)=1\bigr].
    \label{eq:pstar}
\end{equation}
We measure how sufficient a state is by $p^*(x)$, because the decision to answer or gather more information hinges on how likely an answer is to be correct. Sufficiency therefore depends on the model as well as the information. A state is sufficient when its information, together with what the model already knows, lets the model reach the correct answer, whether or not the information itself is complete. The same state can thus be sufficient for one model and insufficient for another.

\begin{figure*}[t]
  \centering
  \includegraphics[width=0.8\textwidth]{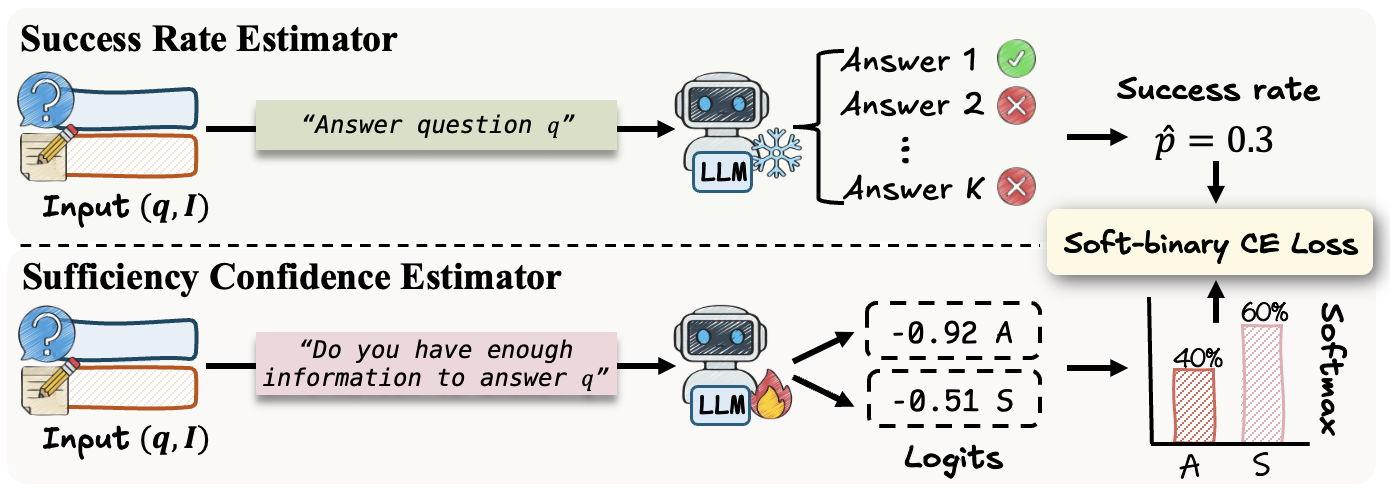}
  \caption{Overview of \ours{}. Each evidence state \(x=(q,I)\) pairs an input question \(q\) with the available evidence \(I\). \textbf{Top:} A frozen answering policy samples \(K\) answers per evidence state, and the fraction graded correct is the empirical success rate \(\hat p(x)\). \textbf{Bottom:} The estimator reads its confidence from the softmax over the logits of \texttt{A} (Answer) and \texttt{S} (Seek), and is trained with soft-label binary cross-entropy to match \(\hat p(x)\).}
  \label{fig:main}
\end{figure*}

\subsection{Evidence States}
\label{sec:info_states}

For each problem $i$, we construct $m_i$ evidence states $x_{i,s}=(q_i,I_{i,s})$, $s=1,\ldots,m_i$, each pairing the question with a subset $I_{i,s}\subseteq E_i$ of its evidence. We write $x$ for a generic state when the indices are not needed. All states of a problem share the question and gold answer and differ only in the evidence shown, so differences in success between them are attributable to the evidence. The model is never told which state it is in and sees only the question and the evidence, so it must judge from their content whether the evidence is enough. Training on the same problem under different evidence teaches it which evidence makes a correct answer reachable.

Evidence states vary in both the amount and the composition of evidence. Since success depends on which evidence is present, not on the order it was gathered, states can combine evidence freely, including states that hold the amount fixed and vary the composition (\S\ref{sec:info_state_analysis}). The states are specific to each dataset (\S\ref{sec:datasets}).

\paragraph{Evidence-state calibration.}
We seek a confidence $c_\theta(x)\in[0,1]$ that a model with parameters $\theta$ computes from the state $x$ alone, before any answer is generated. Evidence-state calibration requires it to satisfy two criteria, where expectations are taken over the constructed evidence states.

\noindent\phantomsection\label{crit:calibration}\textbf{1. Calibration.} Across all states assigned the same confidence $c$, the average success probability is $c$~\citep{guo2017calibration}:
\begin{equation}
    \mathbb{E}\bigl[p^*(x)\mid c_\theta(x)=c\bigr]=c \quad \forall c\in\operatorname{range}(c_\theta).
    \label{eq:goal}
\end{equation}

\noindent\phantomsection\label{crit:awareness}\textbf{2. Evidence sensitivity.} A change in evidence moves confidence as much as it moves success:
\begin{equation}
    c_\theta(x_{i,t}) - c_\theta(x_{i,s}) \approx p^*(x_{i,t}) - p^*(x_{i,s}).
    \label{eq:within}
\end{equation}
Here $x_{i,s}$ and $x_{i,t}$ denote any two evidence states $s$ and $t$ of the same problem $i$, so the difference in confidence should match the difference in success.

To see why calibration alone is not enough, we split the squared error of a confidence by problem. For problem $i$ with $m_i$ states, let $\bar c_{\theta,i}$ and $\bar p^*_i$ denote the averages of $c_\theta$ and $p^*$ over its states, and for any two of its states $s,t\in\{1,\ldots,m_i\}$, let $\delta_{st}=\bigl(c_\theta(x_{i,t})-c_\theta(x_{i,s})\bigr)-\bigl(p^*(x_{i,t})-p^*(x_{i,s})\bigr)$ be the gap between the change in confidence and the change in success from $s$ to $t$, which is zero when Eq.~\eqref{eq:within} holds exactly.

\begin{resultbox}{Error Decomposition}
\begin{proposition}
\label{prop:decomp}
For any confidence $c_\theta$,
\begin{equation}
\begin{split}
    \mathbb{E}\bigl[(c_\theta-p^*)^2\bigr]
    &= \mathbb{E}_i\bigl[(\bar c_{\theta,i}-\bar p^*_i)^2\bigr] \\
    &\quad + \mathbb{E}_i\Bigl[\tfrac{1}{2m_i^2}\textstyle\sum_{s,t}\delta_{st}^2\Bigr],
\end{split}
\label{eq:decomp}
\end{equation}
where the sum runs over all ordered pairs of states of problem $i$, and each problem is weighted by its number of states.
\end{proposition}
\end{resultbox}

The first term is the error in each problem's average success, and the second is the error within each problem, which is zero exactly when \hyperref[crit:awareness]{criterion~2} holds with equality (proof in Appendix~\ref{app:proof_prop}). Calibration does not constrain the second term. Giving every state its problem's average success satisfies \hyperref[crit:calibration]{criterion~1} exactly but leaves this term at the variance of success within each problem, which only following the evidence can reduce. Since both terms are nonnegative, we seek a confidence that drives the left side to zero and thus satisfies both criteria.

\subsection{Monte Carlo Estimation of Success Probability}
\label{sec:success_labels}

Because computing the true success probability $p^*(x)$ exactly would require grading every answer the policy could generate, we estimate it by sampling from the fixed answering policy (Figure~\ref{fig:main}, top). A single answer is only correct or incorrect, so we draw $K$ independent answers per state and average their grades. For each evidence state $x$ of problem $i$, we sample $\hat y^{(1)},\ldots,\hat y^{(K)}\sim\pi_0(\cdot\mid x)$, grade each as $r^{(k)}=g(\hat y^{(k)},y^*_i)$, and define the empirical success rate as
\begin{equation}
    \hat p(x)=\frac{1}{K}\sum_{k=1}^K r^{(k)}.
    \label{eq:phat}
\end{equation}
Since $r^{(k)}\sim\mathrm{Bernoulli}(p^*(x))$, $\mathbb{E}[\hat p(x)\mid x]=p^*(x)$,
so $\hat p(x)$ is an unbiased Monte Carlo estimate of the frozen policy's success probability.

The estimate $\hat p(x)$ is specific to the answering policy $\pi_0$, so the sufficiency of a state $x$ always refers to whether $\pi_0$ can answer correctly given the information in $x$, and the confidence we train is calibrated to this policy. Since computing $\hat p(x)$ requires the gold answer, which is unavailable at decision time, we use it only as a training target.

\subsection{Sufficiency Confidence Estimator}
\label{sec:estimator}

At decision time, the model must therefore predict $\hat p(x)$ from the evidence state alone. We read this prediction from the model's own next-token distribution rather than adding a separate classifier head (Figure~\ref{fig:main}, bottom). Each evidence state is placed in a prompt that asks the model, without answering the task, whether the currently available information is sufficient for it to answer correctly, and defines two actions, \tok{A} to answer now because the information is sufficient, and \tok{S} to seek more information because it is not. Rather than letting the model generate a response, we read the probabilities it assigns to \tok{A} and \tok{S} as the first token of its reply (prompt in Appendix~\ref{app:prompt_estimator}).

Let $\ell_A(x)$ and $\ell_S(x)$ denote the logits assigned to \tok{A} and \tok{S} at the first response position from a single forward pass. We take the softmax over these two logits, giving \(c_\theta(x)=P_\theta(A\mid x)=\exp(\ell_A)/\bigl(\exp(\ell_A)+\exp(\ell_S)\bigr)\). We call \(c_\theta(x)\) the \emph{sufficiency confidence} of state \(x\), and the model that produces it the \emph{sufficiency confidence estimator}.

We train $c_\theta(x)$ to match the empirical success rate using soft-label binary cross-entropy:
\begin{equation}
\begin{split}
    \mathcal L(x)
    ={}&
    -\hat p(x)\log c_\theta(x) \\
    &-
    \bigl(1-\hat p(x)\bigr)
    \log\bigl(1-c_\theta(x)\bigr).
\end{split}
\label{eq:loss}
\end{equation}

Binary cross-entropy is a strictly proper scoring rule~\citep{gneiting2007strictly} and is linear in the target, so with the unbiased labels of \S\ref{sec:success_labels} the expected loss at each state is minimized by $c_\theta(x)=p^*(x)$ despite sampling noise in $\hat p(x)$. Training on this loss also controls both terms of Proposition~\ref{prop:decomp}.

\begin{corollary}
\label{cor:bound}
For any confidence $c_\theta$, let $\mathbb{E}[\mathcal L]$ be the loss of Eq.~\eqref{eq:loss} averaged over all evidence states and over the $K$ sampled answers that determine $\hat p(x)$, and $\mathcal L^*$ the minimum expected loss over all confidences, reached exactly at $c_\theta=p^*$. Then
\begin{equation}
    \mathbb{E}\bigl[(c_\theta-p^*)^2\bigr]\leq\tfrac12\bigl(\mathbb{E}[\mathcal L]-\mathcal L^*\bigr),
    \label{eq:bound}
\end{equation}
so each term of Eq.~\eqref{eq:decomp} is at most $\tfrac12\bigl(\mathbb{E}[\mathcal L]-\mathcal L^*\bigr)$.
\end{corollary}

\noindent Lowering the loss therefore lowers an upper bound on both terms, and the bound reaches zero exactly at $c_\theta=p^*$ (proof in Appendix~\ref{app:proof_cor}).

%% file: sections/experimental_setup.tex
\section{Experimental Setup}
\label{sec:setup}
\paragraph{Datasets.}\label{sec:datasets} We evaluate on two clinical diagnosis datasets and one multi-hop question answering dataset, which range from mostly answered correctly to mostly not (details in Appendix~\ref{app:datasets}). MIMIC~\citep{hager2024evaluation} consists of real emergency-department admissions for four abdominal diseases, and the model chooses among the four. MedQA~\citep{jin2021disease} contains questions from the United States Medical Licensing Examination, and the model names the diagnosis in free text. Both are written as structured clinical cases following AgentClinic~\citep{schmidgall2026agentclinic}. MuSiQue~\citep{trivedi-etal-2022-musique} composes multi-hop questions over Wikipedia from single-hop questions, each answered by one supporting paragraph, and the model gives a short answer. All states of a problem are assigned to the same split to avoid leakage between training and test.

\paragraph{Evidence states.} Clinical cases are divided into history, physical examination, imaging, and other tests such as laboratory and pathology. Each state adds a different subset of the remaining sections to the history.

\begin{table*}[t]
  \centering
  \small
  \setlength{\tabcolsep}{2pt}
  \renewcommand{\arraystretch}{0.9}
  \renewcommand{\sd}[1]{\,{\tiny$\pm$#1}}
  \begin{tabularx}{\textwidth}{@{}l*{9}{>{\centering\arraybackslash}X}@{}}
    \toprule
    & \multicolumn{3}{c}{\textbf{\textit{MIMIC}}} & \multicolumn{3}{c}{\textbf{\textit{MedQA}}} & \multicolumn{3}{c}{\textbf{\textit{MuSiQue}}} \\
    \cmidrule(lr){2-4} \cmidrule(lr){5-7} \cmidrule(l){8-10}
    Method & Brier $\downarrow$ & AUROC $\uparrow$ & $\rho_\Delta$ $\uparrow$ & Brier $\downarrow$ & AUROC $\uparrow$ & $\rho_\Delta$ $\uparrow$ & Brier $\downarrow$ & AUROC $\uparrow$ & $\rho_\Delta$ $\uparrow$ \\
    \midrule
    \multicolumn{10}{@{}l}{\textit{Kimi K2.6}} \\
    Base logit & 0.219\sd{0.000} & 0.686\sd{0.000} & 0.320\sd{0.000} & 0.223\sd{0.000} & 0.707\sd{0.000} & 0.413\sd{0.000} & 0.184\sd{0.000} & 0.750\sd{0.000} & 0.430\sd{0.000} \\
    Verbalized & 0.131\sd{0.001} & 0.686\sd{0.003} & 0.247\sd{0.013} & 0.162\sd{0.001} & 0.650\sd{0.001} & 0.374\sd{0.010} & 0.174\sd{0.003} & 0.724\sd{0.003} & 0.412\sd{0.011} \\
    Platt      & 0.124\sd{0.000} & 0.686\sd{0.000} & 0.296\sd{0.000} & 0.150\sd{0.000} & 0.707\sd{0.000} & 0.438\sd{0.000} & 0.142\sd{0.000} & 0.750\sd{0.000} & 0.431\sd{0.000} \\
    \rowcolor{oursrow}\ours{}    & \bestm{0.105}{0.002} & \bestm{0.844}{0.005} & \bestm{0.474}{0.045} & \bestm{0.141}{0.005} & \bestm{0.795}{0.033} & \bestm{0.467}{0.064} & \bestm{0.134}{0.004} & \bestm{0.787}{0.005} & \bestm{0.632}{0.009} \\
    \midrule
    \multicolumn{10}{@{}l}{\textit{Qwen3.8-27B}} \\
    Base logit & 0.127\sd{0.000} & 0.788\sd{0.000} & 0.353\sd{0.000} & 0.143\sd{0.000} & 0.744\sd{0.000} & 0.413\sd{0.000} & 0.182\sd{0.000} & 0.738\sd{0.000} & 0.274\sd{0.000} \\
    Verbalized & 0.112\sd{0.000} & 0.770\sd{0.007} & 0.222\sd{0.003} & 0.174\sd{0.001} & 0.772\sd{0.001} & 0.355\sd{0.000} & 0.203\sd{0.001} & 0.682\sd{0.002} & 0.345\sd{0.008} \\
    Platt      & 0.099\sd{0.000} & 0.788\sd{0.000} & 0.286\sd{0.000} & 0.126\sd{0.000} & 0.744\sd{0.000} & 0.438\sd{0.000} & 0.123\sd{0.000} & 0.738\sd{0.000} & 0.423\sd{0.000} \\
    \rowcolor{oursrow}\ours{}       & \bestm{0.080}{0.004} & \bestm{0.857}{0.012} & \bestm{0.447}{0.040} & \bestm{0.091}{0.005} & \bestm{0.826}{0.008} & \bestm{0.489}{0.019} & \bestm{0.092}{0.008} & \bestm{0.830}{0.014} & \bestm{0.717}{0.024} \\
    \midrule
    \multicolumn{10}{@{}l}{\textit{Qwen3.5-9B}} \\
    Base logit & 0.124\sd{0.000} & 0.689\sd{0.000} & 0.247\sd{0.000} & 0.268\sd{0.000} & 0.746\sd{0.000} & 0.367\sd{0.001} & 0.197\sd{0.000} & 0.752\sd{0.000} & 0.356\sd{0.000} \\
    Verbalized & 0.143\sd{0.003} & 0.727\sd{0.002} & 0.177\sd{0.011} & 0.290\sd{0.001} & 0.706\sd{0.002} & 0.356\sd{0.004} & 0.145\sd{0.003} & 0.688\sd{0.004} & 0.310\sd{0.012} \\
    Platt      & 0.110\sd{0.000} & 0.689\sd{0.000} & 0.254\sd{0.000} & 0.134\sd{0.000} & 0.746\sd{0.000} & 0.514\sd{0.001} & 0.099\sd{0.000} & 0.752\sd{0.000} & 0.389\sd{0.000} \\
    \rowcolor{oursrow}\ours{}       & \bestm{0.098}{0.010} & \bestm{0.819}{0.028} & \bestm{0.393}{0.034} & \bestm{0.107}{0.005} & \bestm{0.815}{0.011} & \bestm{0.630}{0.012} & \bestm{0.078}{0.002} & \bestm{0.828}{0.006} & \bestm{0.666}{0.015} \\
    \midrule
    \multicolumn{10}{@{}l}{\textit{Qwen3.5-4B}} \\
    Base logit & 0.143\sd{0.000} & 0.761\sd{0.000} & 0.311\sd{0.000} & 0.132\sd{0.000} & 0.763\sd{0.000} & 0.452\sd{0.000} & 0.122\sd{0.000} & 0.779\sd{0.000} & 0.335\sd{0.000} \\
    Verbalized & 0.190\sd{0.001} & 0.763\sd{0.001} & 0.095\sd{0.009} & 0.250\sd{0.005} & 0.739\sd{0.006} & 0.286\sd{0.027} & 0.164\sd{0.000} & 0.601\sd{0.001} & 0.224\sd{0.003} \\
    Platt      & 0.097\sd{0.000} & 0.761\sd{0.000} & 0.274\sd{0.000} & 0.108\sd{0.000} & 0.763\sd{0.000} & 0.488\sd{0.000} & 0.097\sd{0.000} & 0.779\sd{0.000} & 0.400\sd{0.000} \\
    \rowcolor{oursrow}\ours{}       & \bestm{0.085}{0.003} & \bestm{0.830}{0.014} & \bestm{0.458}{0.097} & \bestm{0.103}{0.003} & \bestm{0.773}{0.011} & \bestm{0.524}{0.072} & \bestm{0.086}{0.004} & \bestm{0.818}{0.007} & \bestm{0.606}{0.020} \\
    \bottomrule
  \end{tabularx}
  \caption{Results on the three datasets for Kimi K2.6, Qwen3.8-27B, Qwen3.5-9B, and Qwen3.5-4B, comparing our method with baselines. Results are mean $\pm$ SD over three seeds. The best result for each model is in bold.}
  \label{tab:main_results}
\end{table*}

\begin{figure}[!ht]
  \centering
  \includegraphics[width=0.9\columnwidth]{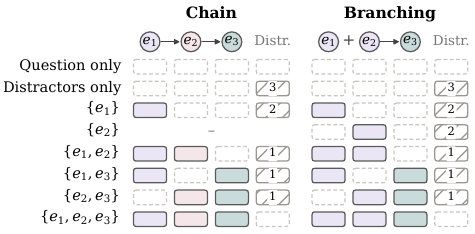}
  \caption{Evidence states of MuSiQue for chain and branching questions. Hatched cells give the number of distractors padding each state to three paragraphs.}
  \label{fig:musique_states}
\end{figure}

In MuSiQue, we use 3-hop questions, whose evidence units are the three supporting paragraphs, one per sub-question. In chain questions, each answer is needed to ask the next sub-question (\(e_1 \rightarrow e_2 \rightarrow e_3\)), while in branching questions, \(e_1\) and \(e_2\) answer independent sub-questions that both feed the final one, \(e_3\). States range from the question alone and distractors only (related paragraphs that MuSiQue retrieves but that do not help answer the question) to one, two, or all three supporting paragraphs (Figure~\ref{fig:musique_states}, examples in Appendix~\ref{app:examples}). Every state but the question-only one shows three paragraphs, padded with distractors in one shuffled order per question, so neither the number nor the position of paragraphs reveals the state.

\paragraph{Baselines.} We compare against three baselines on the same untrained model. \emph{Base logit} (\emph{Base}) is our sufficiency confidence estimator before training, reading \(P(A\mid x)\) under the prompt of \S\ref{sec:estimator}. \emph{Verbalized} asks the model to state the probability that it could answer correctly from the available information~\citep{tian-etal-2023-just}. \emph{Platt} scaling~\citep{platt1999probabilistic} rescales the base logit to fit the success rates in the training set, so it can recalibrate the confidence but not reorder states (prompts in Appendix~\ref{app:prompts}).

\paragraph{Implementation details.} We evaluate four answering models, three Qwen models of increasing size (Qwen3.5-4B, Qwen3.5-9B, and Qwen3.8-27B) and Kimi K2.6 from a different model family. The answering policy \(\pi_0\) can be any frozen policy. Our main experiments use the estimator's own base model, to study how well a model can judge its own evidence state. Success rates are estimated from \(K=10\) sampled answers per state. Further details are in Appendix~\ref{app:training}.

\paragraph{Metrics.}\label{sec:metrics} Let \(c_{i,s}=c_\theta(x_{i,s})\) be the sufficiency confidence of state \(s\) of problem \(i\) and \(\hat p_{i,s}\) its empirical success rate. The \emph{Brier} score, the mean of \((c_{i,s}-\hat p_{i,s})^2\) over the test set, measures overall error. \emph{AUROC}, averaged over the \(K\) sampled answers, measures how well confidence ranks correct answers above incorrect ones. Both pool the whole test set, so they reward knowing which problems are hard. The \emph{change correlation} \(\rho_\Delta=\operatorname{Corr}(c_{i,t}-c_{i,s},\,\hat p_{i,t}-\hat p_{i,s})\), computed over all pairs of states of the same problem from a state \(s\) to a state \(t\) with more evidence (Appendix~\ref{app:info_state_analysis}), instead tests whether confidence follows success as the evidence state changes, as \hyperref[crit:awareness]{criterion~2} requires.

%% file: sections/results.tex
\section{Results}
\label{sec:results}
\subsection{Overall Performance}
\label{sec:overall}

\ours{} achieves the best result on every metric in every setting (Table~\ref{tab:main_results}). Each model is calibrated to its own success rates, so methods are compared only for the same model (Appendix~\ref{app:policies}). \ours{} lowers the Brier score by 20--60\% relative to the base logit and by 5--28\% relative to Platt scaling, the strongest baseline. \ours{} not only calibrates better but also ranks states better. Platt scaling fits the same labels but only rescales the base confidence, leaving the order of states, and thus AUROC and selective accuracy, unchanged, whereas \ours{} raises AUROC by up to 0.16. \ours{} also raises the change correlation \(\rho_\Delta\) above every baseline, so its confidence moves with success as the evidence changes, which \S\ref{sec:info_state_analysis} examines in detail. 

\begin{figure}[!ht]
  \centering
  \includegraphics[width=0.85\columnwidth]{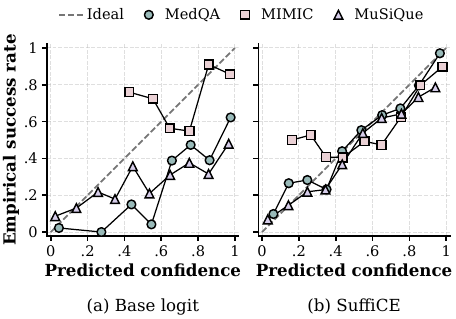}
  \caption{\ours{} brings confidence closer to perfect calibration on every dataset. Calibration curves on the test set (a) before and (b) after \ours{}.}
  \label{fig:overall}
\end{figure}

\paragraph{Calibration across datasets.} The base logit is miscalibrated by different amounts across datasets, and \ours{} improves calibration on all three, moving each curve toward the ideal diagonal (Figure~\ref{fig:overall}, Qwen3.5-9B). The other models follow the same pattern (Appendix~\ref{app:reliability}).

\paragraph{Selective answering.} If the model answers only the questions it is most confident about and abstains on the rest, \ours{} is the most accurate at nearly every percentage of questions answered (Table~\ref{tab:selective} and Appendix~\ref{app:coverage}). On MIMIC, the questions it is most confident about are almost always answered correctly.

\begin{table}[!ht]
  \centering
  \small
  \setlength{\tabcolsep}{5pt}
  \renewcommand{\arraystretch}{0.9}
  \begin{tabular}{@{}llccc@{}}
    \toprule
    & Answered & Base & Verb. & \ours{} \\
    \midrule
    MIMIC   & 10\% & 97.2 & 95.3 & \textbf{100.0} \\
            & 30\% & 93.4 & 95.3 & \textbf{98.5} \\
            & 50\% & 89.9 & 92.4 & \textbf{95.5} \\
    \midrule
    MedQA   & 10\% & 93.7 & 74.3 & \textbf{99.1} \\
            & 30\% & 77.8 & 73.4 & \textbf{84.2} \\
            & 50\% & 65.4 & 63.3 & \textbf{74.0} \\
    \midrule
    MuSiQue & 10\% & 55.2 & 52.5 & \textbf{68.9} \\
            & 30\% & 45.3 & 47.7 & \textbf{56.2} \\
            & 50\% & 39.4 & 34.7 & \textbf{42.6} \\
    \bottomrule
  \end{tabular}
  \caption{Accuracy (\%) when the model answers the given percentage of questions it is most confident about.}
  \label{tab:selective}
\end{table}

\subsection{Ablation}
\label{sec:ablation}

\begin{figure}[!ht]
  \centering
  \includegraphics[width=0.95\columnwidth]{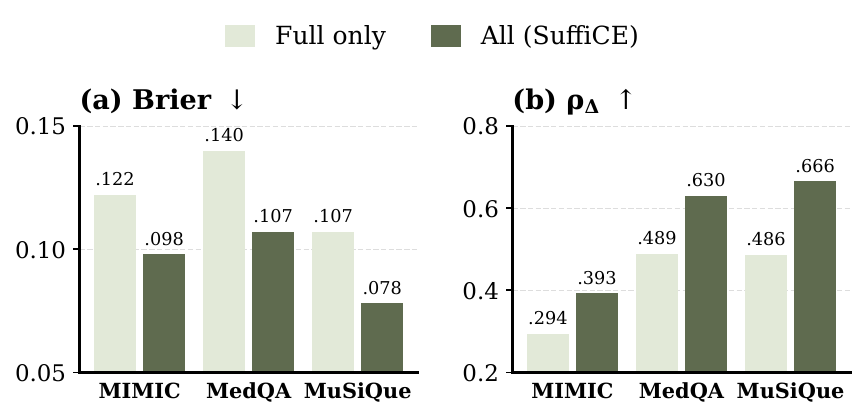}
  \caption{Training across evidence states improves both calibration and evidence sensitivity. Estimators trained for the same number of epochs on full states only versus on all evidence states (\ours{}).}
  \label{fig:ablation}
\end{figure}

To test whether the estimator needs diverse evidence states, we train Qwen3.5-9B only on the full state of each training problem and evaluate it on the whole test set (Figure~\ref{fig:ablation}). Success on full states still varies across problems, so this estimator learns which problems are hard but never sees missing evidence. \ours{} outperforms it on every metric and dataset, both with the same number of training epochs and with the same number of steps (Appendix~\ref{app:ablation_steps}).
\subsection{Evidence-State Analysis}
\label{sec:info_state_analysis}

\paragraph{Confidence rises with the decisive evidence, not with more evidence.} \ours{} closely tracks the empirical success rate across states, rising only when new evidence makes the question answerable, whereas the base logit rises with every paragraph added (Figure~\ref{fig:musique_evidence}, more measures in Appendix~\ref{app:info_state_analysis}). For both question types, success stays near zero until the final hop \(e_3\) is shown. On chain questions, the base logit treats each earlier hop as progress and is confident enough to answer given \(\{e_1,e_2\}\) (0.71), although the model almost never succeeds (0.06), while \ours{} stays low (0.16) until \(e_3\) arrives. On branching questions, the base logit rises with each branch, while \ours{} stays low for either branch alone and for both, since neither helps without \(e_3\). \ours{} is never told the question type or which paragraph fills which hop, so it judges from the content alone whether the reasoning chain is complete. It thus separates evidence that makes a question answerable from evidence that only adds information (Appendix~\ref{app:case_study}).

\begin{figure}[!ht]
  \centering
  \includegraphics[width=\columnwidth]{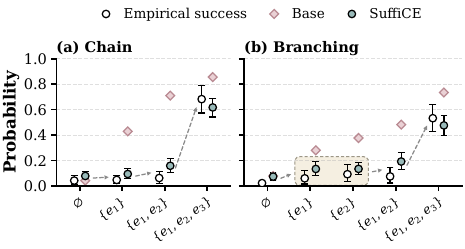}
  \caption{\ours{} tracks the empirical success rate as supporting paragraphs are added, while the base logit rises with every paragraph. MuSiQue test set, for (a) chain and (b) branching questions.}
  \label{fig:musique_evidence}
\end{figure} 

\begin{figure}[!ht]
  \centering
  \includegraphics[width=\columnwidth]{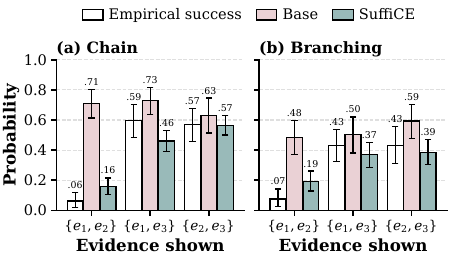}
  \caption{With the same amount of evidence, \ours{} tells which states can be answered. Each MuSiQue state shows two of the three supporting paragraphs, for (a) chain and (b) branching questions. Only \ours{} stays low until the final hop \(e_3\) is shown.}
  \label{fig:musique_equal_count}
\end{figure}

\paragraph{Holding the amount fixed, content decides success.} Even when every state has the same amount of evidence, \ours{} distinguishes which states can be answered (Figure~\ref{fig:musique_equal_count}). Each state shows two of the three supporting paragraphs, and success is high only when the final hop \(e_3\) is among them. The base logit is similarly confident on all three states, responding to how much evidence there is, whereas \ours{} stays low on \(\{e_1,e_2\}\) and rises for both states with \(e_3\).

\paragraph{Confidence separates decisive from suggestive evidence.} In clinical diagnosis, the decisive evidence differs from patient to patient. Table~\ref{tab:case_study_clinical} shows a MedQA case of Hirschsprung disease. The X-ray is the pivot of the workup. It reveals a distal obstruction and prompts the rectal biopsy, but it cannot tell Hirschsprung disease from other causes of obstruction. Only the biopsy, which finds no ganglion cells, confirms the diagnosis. The base logit is nearly certain in every state, whereas \ours{} stays uncertain with the X-ray in state \(\{1,4\}\) and becomes confident only once the biopsy is added in state \(\{1,3,4\}\). With a suitable confidence threshold, \ours{} would therefore seek more information before answering. Unlike the base logit, \ours{} thus recognizes which evidence settles the diagnosis and which only signals that more information is needed.

\begin{table}[!ht]
  \centering
  \small
  \renewcommand{\arraystretch}{1.1}
  \begin{tabularx}{\columnwidth}{@{}l@{\hspace{5pt}}>{\raggedright\arraybackslash}X@{}}
    \toprule
    \multicolumn{2}{@{}>{\raggedright\arraybackslash}p{\columnwidth}@{}}{\textit{A 2-day-old boy has bilious vomiting, has not passed meconium since birth, and has a distended abdomen. What is the diagnosis?} \textbf{Answer:} Hirschsprung disease} \\
    \midrule
    \cnum{1}{histstrong}~\textsc{History} & Home birth without complications, bilious vomiting, and no meconium since birth. \\
    \cnum{2}{examstrong}~\textsc{PE} & Firm, distended abdomen with decreased bowel sounds and palpable bowel loops. \\
    \cnum{3}{nonimgstrong}~\textsc{Biopsy} & Rectal biopsy: no ganglion cells in the submucosa or myenteric plexus. \\
    \cnum{4}{imgstrong}~\textsc{Imaging} & Abdominal X-ray: dilated bowel loops and no rectal gas, suggesting a distal obstruction. \\
    \bottomrule
  \end{tabularx}

  \vspace{3pt}
  \renewcommand{\arraystretch}{1.0}
  \setlength{\tabcolsep}{1.2pt}
  \begin{tabular*}{\columnwidth}{@{}l@{\extracolsep{\fill}}ccccc@{}}
    \toprule
    State & {\footnotesize$\{1\}$} & {\footnotesize$\{1,\!2\}$} & \hlnode{hlmedT}{\footnotesize$\{1,\!4\}$} & {\footnotesize$\{1,\!3,\!4\}$} & {\footnotesize$\{1,\!2,\!3,\!4\}$}  \\
    \midrule
    Success & 0.20 & 0.40 & 0.40 & 1.00 & 1.00  \\
    Base     & 0.96 & 0.99 & 0.99 & 1.00 & 1.00  \\
    \ours{}  & 0.50 & 0.47 & \hlnode{hlmedB}{0.51} & 0.97 & 0.97  \\
    \bottomrule
  \end{tabular*}
  \hlcolumn[1.5pt]{hlmed}{15pt}%
  \caption{A MedQA case of Hirschsprung disease (Qwen3.5-9B). Each state is the set of evidence shown. Success is the fraction of ten sampled answers that are correct. Base and \ours{} give the sufficiency confidence.}
  \label{tab:case_study_clinical}
\end{table}

\subsection{Cross-Policy Confidence Estimation}
\label{sec:cross_policy}

\begin{figure}[!ht]
  \centering
  \includegraphics[width=\columnwidth]{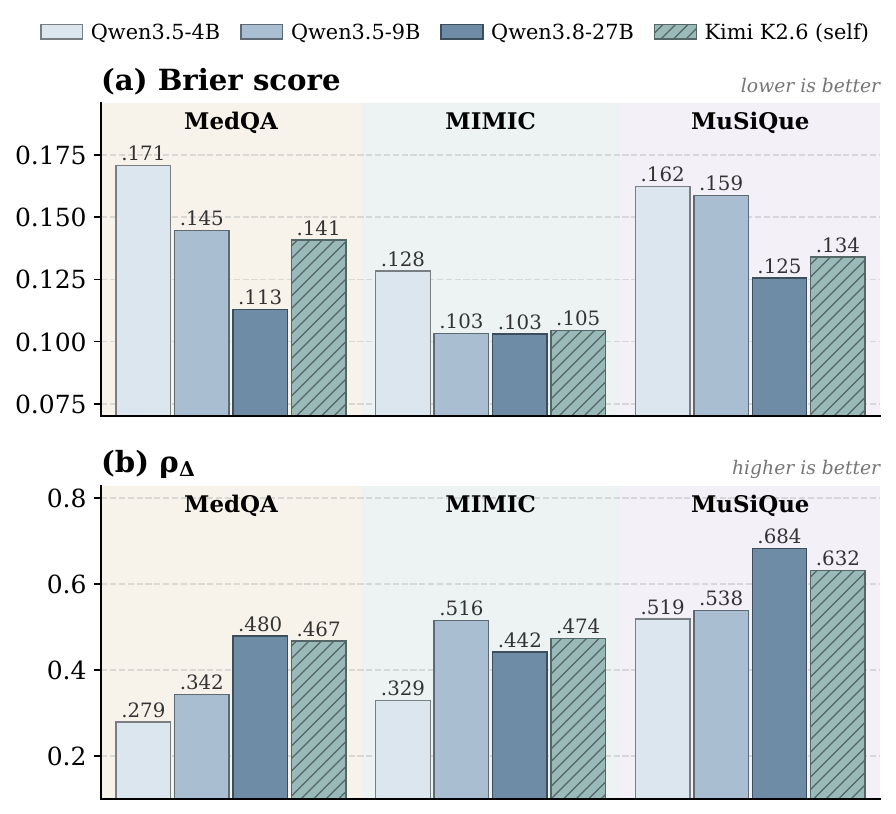}
  \caption{Qwen estimators of three sizes are trained to predict the success rates of Kimi K2.6, compared with Kimi K2.6 predicting its own (self).}
  \label{fig:cross_policy}
\end{figure}

To test whether the estimator must be the answering model, we fix the answering policy to Kimi K2.6 and train Qwen estimators of three sizes on its success rates (Figure~\ref{fig:cross_policy}). A large enough estimator predicts another model's success as well as that model does itself. The 27B estimator matches or exceeds Kimi K2.6's own estimate on every dataset in Brier score, and in \(\rho_\Delta\) on MedQA and MuSiQue. Trained only on Kimi K2.6's success rates and never shown its answers, the estimator thus learns to anticipate from the evidence state alone how well another model can answer. Smaller estimators are less accurate, and the 4B estimator falls well short of Kimi K2.6's own estimate.

%% file: sections/conclusion.tex
\section{Conclusion}
\label{sec:conclusion}

To our knowledge, this is the first work to train a model's confidence before answering to be calibrated across varying evidence states. The decision to answer or gather more evidence is made before any answer exists, so a model is only as reliable as this judgment. We formalized it as \emph{evidence-state calibration}, requiring confidence to be calibrated and to change as much as the chance of success whenever the evidence changes, which calibration alone does not guarantee. \ours{} outperforms every baseline, and its confidence rises with the decisive evidence, not with every piece added. Such confidence is a step toward language models that can judge, in high-stakes settings, when the evidence at hand is enough to answer.

%% file: sections/appendix.tex
\clearpage
\section{Experimental Details}
\label{app:experimental}

\subsection{Datasets}
\label{app:datasets}

The three datasets range widely in difficulty (Table~\ref{tab:datasets}). For Qwen3.5-9B, the mean success rate on the test set is 0.81 on MIMIC, 0.50 on MedQA, and 0.25 on MuSiQue, so we test the confidence both where most states are answered correctly and where most are not.

\begin{table}[!ht]
  \centering
  \small
  \setlength{\tabcolsep}{4pt}
  \begin{tabular}{@{}lccc@{}}
    \toprule
    & MIMIC & MedQA & MuSiQue \\
    \midrule
    Answer & 4-way choice & free text & short answer \\
    Problems & 140 / 60 & 146 / 63 & 200 / 80 \\
    States & 840 / 360 & 728 / 302 & 1,500 / 600 \\
    \midrule
    \multicolumn{4}{@{}l}{\textit{Mean success rate \(\hat p\)}} \\
    \quad Qwen3.5-4B  & 0.82 / 0.81 & 0.45 / 0.44 & 0.27 / 0.25 \\
    \quad Qwen3.5-9B  & 0.84 / 0.81 & 0.50 / 0.50 & 0.29 / 0.25 \\
    \quad Qwen3.8-27B & 0.87 / 0.83 & 0.64 / 0.62 & 0.34 / 0.31 \\
    \quad Kimi K2.6   & 0.87 / 0.82 & 0.76 / 0.71 & 0.44 / 0.43 \\
    \bottomrule
  \end{tabular}
  \caption{Dataset statistics, given as train / test. The mean success rate \(\hat p\) of each answering model is taken over all evidence states.}
  \label{tab:datasets}
\end{table}

\paragraph{MIMIC.} We use emergency admissions from MIMIC-IV-Ext Clinical Decision Making~\citep{hager2024evaluation}. We sample 50 admissions for each of four abdominal diseases, appendicitis, cholecystitis, diverticulitis, and pancreatitis, and the task is to choose one of the four. We convert each admission to the OSCE layout of AgentClinic~\citep{schmidgall2026agentclinic}. The history, laboratory results, and radiology reports are copied verbatim, and an LLM (GPT-5.1) writes only a one-sentence objective naming the presenting complaints and splits the free-text physical examination into sections without changing its values. Microbiology results are never shown, since cultures return after the diagnostic decision. We split admissions into training and test sets stratified by disease.

\paragraph{MedQA.} We use the OSCE-style cases of the extended AgentClinic-MedQA set~\citep{schmidgall2026agentclinic}, which are derived from USMLE questions~\citep{jin2021disease}, and the task is to name the diagnosis in free text. We exclude 5 of its 214 cases, four with no test results and one with a nonstandard format, which leaves 209. States that show only imaging or only the other tests exist for the 97 cases that have both kinds of test. We split cases by diagnosis, which keeps near-duplicate vignettes on the same side and makes every test diagnosis unseen during training.

\paragraph{MuSiQue.} We use 3-hop questions from MuSiQue-Ans~\citep{trivedi-etal-2022-musique}, and the task is to give a short answer. For each question, we sample three distractors from its non-supporting paragraphs. Its three supporting paragraphs and three distractors are shuffled once with a seed derived from the question, and every state shows its paragraphs in that order. Training questions come from the MuSiQue training set and test questions from its development set, since its test set has no released answers, with no single-hop question shared between them. MuSiQue's own decompositions distinguish two question types (Table~\ref{tab:musique_types}). In a chain question each hop needs the answer of the previous one, whereas in a branching question the first two hops are independent and only the final hop combines them. A branching question therefore has a state with \(e_2\) alone, while in a chain \(e_2\) is only useful after \(e_1\). The full inputs of the branching example appear in Table~\ref{tab:full_musique}.

\begin{table}[h]
  \centering
  \small
  \renewcommand{\arraystretch}{1.15}
  \begin{tabularx}{\columnwidth}{@{}c@{\hspace{6pt}}>{\raggedright\arraybackslash}X@{\hspace{6pt}}>{\raggedright\arraybackslash}p{0.34\columnwidth}@{}}
    \toprule
    \multicolumn{3}{@{}c@{}}{\textbf{Chain (3hop1)}\quad \enodesm{1}{hoplight}\(\,\rightarrow\,\)\enodesm{2}{nonimglight}\(\,\rightarrow\,\)\enodesm{3}{finallight}} \\
    \addlinespace[2pt]
    \multicolumn{3}{@{}>{\raggedright\arraybackslash}p{\columnwidth}@{}}{\textit{Who was appointed arbitrator by the first convert to the faith that is the most common religion in Southeast Asia?}} \\
    \multicolumn{3}{@{}l}{\textbf{Answer:} Abu Musa Ashaari} \\
    \midrule
    Hop & Sub-question & Answer \\
    \midrule
    \enodesm{1}{hoplight}    & What is the most common religion in Southeast Asia? & Islam \\
    \enodesm{2}{nonimglight} & Who was the first convert to the faith called \textbf{\#1}? & Ali \\
    \enodesm{3}{finallight}  & Who was appointed arbitrator by \textbf{\#2}? & Abu Musa Ashaari \\
    \bottomrule
  \end{tabularx}

  \vspace{8pt}
  \begin{tabularx}{\columnwidth}{@{}c@{\hspace{6pt}}>{\raggedright\arraybackslash}X@{\hspace{6pt}}>{\raggedright\arraybackslash}p{0.34\columnwidth}@{}}
    \toprule
    \multicolumn{3}{@{}c@{}}{\textbf{Branching (3hop2)}\quad \enodesm{1}{hoplight}\(\,+\,\)\enodesm{2}{hoplight}\(\,\rightarrow\,\)\enodesm{3}{finallight}} \\
    \addlinespace[2pt]
    \multicolumn{3}{@{}>{\raggedright\arraybackslash}p{\columnwidth}@{}}{\textit{When did the party Alaska generally supports take control of the determiner of rules of the US House and US Senate?}} \\
    \multicolumn{3}{@{}l}{\textbf{Answer:} January 2015} \\
    \midrule
    Hop & Sub-question & Answer \\
    \midrule
    \enodesm{1}{hoplight}   & Which political party does Alaska generally support? & Republicans \\
    \enodesm{2}{hoplight}   & Who determines the rules of the US House and US Senate? & The Senate \\
    \enodesm{3}{finallight} & When did \textbf{\#1} take control of the \textbf{\#2}? & January 2015 \\
    \bottomrule
  \end{tabularx}
  \caption{The two MuSiQue question types, with sub-questions from MuSiQue's decompositions (capitalization normalized). \textbf{\#}\(k\) stands for the answer of hop \(k\), supported by paragraph \(e_k\). In the chain, each hop depends on the previous one, so each stage has its own color. In the branching question, \(e_1\) and \(e_2\) are independent and share a color, and \(e_3\) combines their answers.}
  \label{tab:musique_types}
\end{table}

\subsection{Answering Policies}
\label{app:policies}

The main experiments study self-calibration. Each model in Table~\ref{tab:main_results} serves as both the answering policy and the estimator, so it estimates its own chance of success, and every method for that model is compared against the exact same targets.

\paragraph{Targets for self-calibration.} For each model, the frozen model with thinking disabled samples \(K=10\) answers per evidence state, and the fraction graded correct is the success rate of that state. These success rates are the only targets used for that model. The base logit reads \(P(A\mid x)\) from this model before training, and verbalized confidence prompts this model for the probability that it could answer correctly. Platt scaling fits its two parameters to this model's success rates on the training set, and \ours{} trains a LoRA adapter on a copy of this model to predict the same success rates. All four are evaluated against the success rates of this model on the test set. Methods are therefore compared only for the same model, since each model has its own targets. The cross-policy experiment of \S\ref{sec:cross_policy} is the only exception, where Qwen estimators are trained on and evaluated against the success rates of Kimi K2.6.

\paragraph{Difficulty.} Success rates differ across answering models, little on MIMIC but widely on MedQA and MuSiQue (Table~\ref{tab:datasets}), so the same evidence state has a different target for each answering policy.

\paragraph{Precision of the targets.} Each target is estimated from \(K=10\) answers, so even a perfect estimator that outputs the true success rate \(p^*\) has a nonzero Brier score against it. This \emph{noise floor} is the expected binomial sampling variance \(p^*(1-p^*)/K\), which we estimate without bias by \(\hat p(1-\hat p)/(K-1)\) averaged over the test set. On every dataset, the Brier score of \ours{} is over 13 times this floor (Table~\ref{tab:noise_floor}). Sampling noise therefore accounts for only a small part of any method's error, and ten answers per state are enough to compare estimators.

\begin{table}[!ht]
  \centering
  \small
  \setlength{\tabcolsep}{5pt}
  \begin{tabular}{@{}lcc@{}}
    \toprule
    Dataset & Noise floor & \ours{} Brier \\
    \midrule
    MIMIC   & 0.003 & 0.092 \\
    MedQA   & 0.008 & 0.111 \\
    MuSiQue & 0.007 & 0.097 \\
    \bottomrule
  \end{tabular}
  \caption{Noise floor, the expected Brier score of a perfect estimator against ten-answer success rates, compared with the Brier score of \ours{} (Table~\ref{tab:main_results}). Both are averaged over the four answering models, on the test set.}
  \label{tab:noise_floor}
\end{table}

\subsection{Training Details}
\label{app:training}

The settings shared by all models and datasets are listed in Table~\ref{tab:hyperparams_shared}. All models are run with thinking disabled. Answers are sampled at temperature 1.0 without top-\(p\) or top-\(k\) truncation. The estimator is a rank-32 LoRA adapter~\citep{hu2022lora} on a copy of the estimator's base model, which is the answering model itself except in the cross-policy experiment (\S\ref{sec:cross_policy}), initialized to have no effect, so before training it gives the base logit. All training and sampling run on Tinker.

\paragraph{Grading.} Each sampled answer is graded as correct or incorrect against the reference. On MIMIC, an answer is correct if, after lowercasing and removing everything but letters, it is exactly the name of the correct disease among the four options, and any other answer is incorrect. On MedQA, GPT-5.6 Luna judges whether a physician would accept the answer as the same diagnosis as the reference. On MuSiQue, an answer is correct if it matches the reference or one of its aliases after SQuAD normalization (lowercasing and removing punctuation and articles), and otherwise GPT-5.6 Luna judges it given the question and the reference. Both judges decode greedily, and their prompts are in Appendix~\ref{app:judges}.

\begin{table}[h]
  \centering
  \small
  \renewcommand{\arraystretch}{0.88}
  \begin{tabular}{@{}ll@{}}
    \toprule
    Setting & Value \\
    \midrule
    \multicolumn{2}{@{}l}{\textit{Success-rate estimation}} \\
    \quad Answers per state \(K\) & 10 \\
    \quad Temperature & 1.0 \\
    \quad Top-\(p\) / top-\(k\) & none \\
    \quad LLM judge (if used) & GPT-5.6 Luna \\
    \midrule
    \multicolumn{2}{@{}l}{\textit{Estimator training}} \\
    \quad LoRA rank & 32 \\
    \quad Loss & soft-label BCE \\
    \quad Optimizer & Adam~\citep{kingma2015adam} \\
    \quad Learning rate & \(1\times10^{-4}\) \\
    \quad Adam \(\beta_1\), \(\beta_2\), \(\epsilon\) & 0.9, 0.95, \(10^{-8}\) \\
    \quad Learning-rate schedule & linear decay to 0 \\
    \quad Batch size & 8 \\
    \quad Seeds & 3 (batch order, adapter init.) \\
    \midrule
    \multicolumn{2}{@{}l}{\textit{Verbalized baseline}} \\
    \quad Runs & 3 \\
    \bottomrule
  \end{tabular}
  \caption{Settings shared by all models and datasets.}
  \label{tab:hyperparams_shared}
\end{table}

\clearpage
\section{Prompts}
\label{app:prompts}

We list every prompt verbatim, grouped by role. Placeholders are in braces. \texttt{\{question\}} is the MuSiQue question or the clinical case's objective, \texttt{\{context\}} and \texttt{\{case\}} are the evidence shown in the current state, and \texttt{\{task line\}} is the dataset-specific line of Table~\ref{tab:task_lines}. All prompts are given as a single user message with thinking disabled, except the judges, which also receive a system message.

\begin{table}[h]
  \centering
  \small
  \begin{tabularx}{\columnwidth}{@{}l>{\raggedright\arraybackslash}X@{}}
    \toprule
    Dataset & \texttt{\{task line\}} \\
    \midrule
    MIMIC & Which of the following is the most likely diagnosis: appendicitis, cholecystitis, diverticulitis, or pancreatitis? \\
    MedQA & What is the most likely diagnosis? \\
    MuSiQue & (none, the question is the whole task) \\
    \bottomrule
  \end{tabularx}
  \caption{The dataset-specific task line of the sufficiency and verbalized prompts.}
  \label{tab:task_lines}
\end{table}

\subsection{Sufficiency Confidence}
\label{app:prompt_estimator}
The sufficiency confidence estimator and the base logit read the probabilities of \texttt{A} and \texttt{S} as the first token of the reply (\S\ref{sec:estimator}).

\begin{promptbox}{Sufficiency prompt (all datasets)}
Decide whether the currently available information is sufficient for you to answer the task correctly. Do not answer the task itself.

A = the information is sufficient; answer now.
S = the information is insufficient; seek more information.

Task:
{question}
{task line}

Available information:
{context}

Is the available information sufficient? Respond with exactly one letter: A or S.
\end{promptbox}

\subsection{Answering}
The frozen answering policy samples ten answers per state with these prompts, from which the empirical success rate is computed (\S\ref{sec:success_labels}).

\begin{promptbox}{Answering prompt: MIMIC}
{case}

This is all the information available about the patient.
Based on this information, which of the following is the most likely diagnosis?
appendicitis
cholecystitis
diverticulitis
pancreatitis
Respond with only one of these four words, nothing else.
\end{promptbox}

\begin{promptbox}{Answering prompt: MedQA}
{case}

This is all the information available about the patient.
Based on this information, what is the most likely diagnosis?
Respond with only the diagnosis, nothing else.
\end{promptbox}

\begin{promptbox}{Answering prompt: MuSiQue}
Answer the question using the provided information and your own knowledge.
If you are unsure, give your best guess.
Give only the final answer, with no explanation, in at most 100 characters.

{case}

Answer:
\end{promptbox}

\subsection{Verbalized Baseline}
The first number in the reply is taken as the confidence, with ``85\%'' read as 0.85.

\begin{promptbox}{Verbalized confidence prompt (all datasets)}
Based on the currently available information, estimate the probability that you could answer the task correctly without seeking any additional information.
Do not answer the task itself.
Respond with a single number between 0 and 1, where 0 means the current information is certainly insufficient for a correct answer and 1 means it is certainly sufficient.
Task:
{question}
{task line}
Available information:
{context}
Probability:
\end{promptbox}

\subsection{LLM Judges}
\label{app:judges}
MedQA answers are graded by the judge below. MuSiQue answers that are not a normalized exact match with the reference or one of its aliases are graded by the MuSiQue judge. MIMIC answers are graded by exact match. All methods for a model are evaluated against the same grades, so differences between methods do not come from differences in grading.

\begin{promptbox}{MedQA judge: system}
You are an expert physician grading a diagnostic answer against a reference diagnosis.
First explain your reasoning briefly, then give a grade of 1 (correct) or 0 (incorrect).
\end{promptbox}

\begin{promptbox}{MedQA judge: user}
Correct diagnosis: {correct_diagnosis}
Doctor's diagnosis: {predicted_diagnosis}

Grade 1 if a physician would accept the doctor's diagnosis as the same diagnosis as the correct one. Accept synonyms, abbreviations, alternate spellings, historical or modern terminology, and diagnoses at an equivalent level of specificity. A more specific form of the correct diagnosis is also acceptable.

Grade 0 if the doctor's diagnosis:
- is a different condition, even a closely related one,
- is only a broader disease category that contains the correct diagnosis but does not identify the more specific diagnosis the correct one requires,
- lists several possibilities without committing to the correct one, or
- is missing.

Respond in exactly this format:
<reasoning>At most three sentences comparing the two diagnoses.</reasoning>
<grade>1 or 0</grade>
\end{promptbox}

\begin{promptbox}{MuSiQue judge: system}
You are grading short answers to multi-hop questions against a reference answer.
First explain your reasoning briefly, then give a grade of 1 (correct) or 0 (incorrect).
\end{promptbox}

\begin{promptbox}{MuSiQue judge: user}
Question: {question}
Reference answer: {answer}
Model answer: {predicted}

Grade 1 only if the response clearly commits to an answer that is semantically correct according to the reference. Grade 0 if it gives multiple alternative answers without clearly committing to the correct one, or if it gives a wrong answer to the question alongside the correct one. Ignore extra details that do not answer the question, even if they are inaccurate. Differences in wording or formatting are acceptable when they preserve the same underlying answer. A broader place, category, or entity, or a related-but-different event, is not equivalent to the reference. Dates are the only exception: differences in date precision are acceptable when they preserve the correct time period, so an exact date, month/year, year, or decade may be treated as equivalent when consistent with the reference. Extremely broad periods such as centuries are not sufficient substitutes for a specific year or date.

Respond in exactly this format:
<reasoning>At most two sentences.</reasoning>
<grade>1 or 0</grade>
\end{promptbox}

\clearpage
\section{Additional Results}
\label{app:additional_results}

\subsection{Calibration Curves for All Models}
\label{app:reliability}

\begin{figure}[H]
  \centering
  \includegraphics[width=\columnwidth]{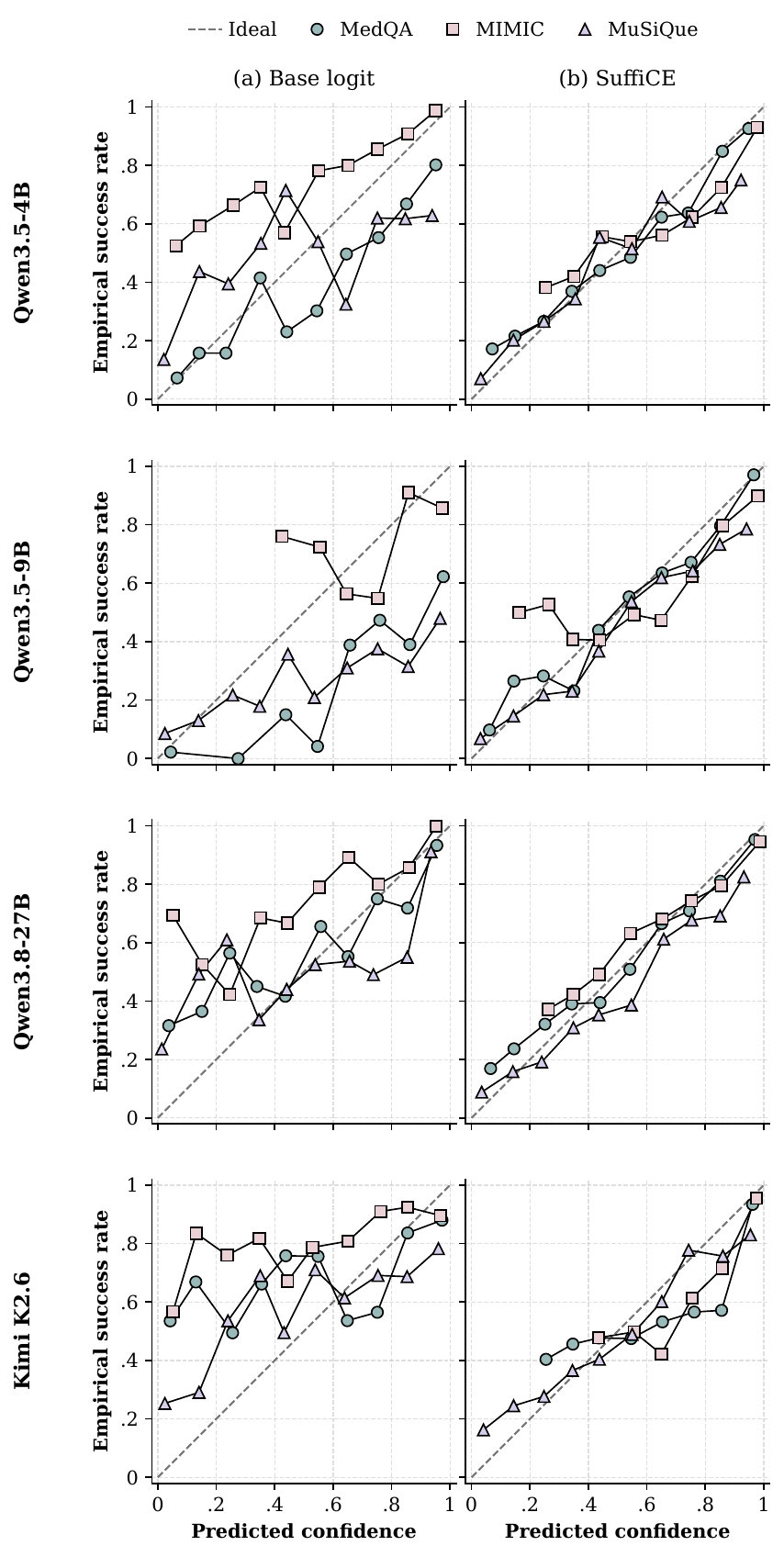}
  \caption{Calibration curves of Qwen3.5-4B, Qwen3.5-9B, Qwen3.8-27B, and Kimi K2.6 on the test set, (a) before and (b) after \ours{}.}
  \label{fig:reliability_grid}
\end{figure}

We extend the calibration curves of Figure~\ref{fig:overall} to all four answering models (Figure~\ref{fig:reliability_grid}).

\paragraph{The base confidence errs in no consistent direction.} Under the same prompt, Qwen3.5-9B is overconfident on every dataset, Qwen3.8-27B and Kimi K2.6 are underconfident on every dataset, and Qwen3.5-4B is overconfident on MedQA but underconfident on MIMIC and MuSiQue. No single correction therefore repairs the base confidence across models and datasets. Many base curves are also not monotone, so a state with higher confidence is not reliably more answerable. Platt scaling can shift such a curve toward the diagonal but cannot straighten it, since it keeps the order of states.

\paragraph{\ours{} moves every curve toward the diagonal.} After training, every Qwen curve lies closer to the diagonal. The confidence also adapts to the range of each dataset, spanning low values on MuSiQue, where most states cannot be answered, and concentrating near the top on MIMIC, where most are answered correctly. The same prompt and objective produce both, so the estimator learns the scale of each dataset from its targets rather than imposing a fixed one.

\paragraph{\ours{} carries over to another model family.} Before training, Kimi K2.6 is strongly underconfident on every dataset. \ours{} moves its curve toward the diagonal on each.

\subsection{Selective Answering for All Models}
\label{app:coverage}

\begin{figure*}[t]
  \centering
  \includegraphics[width=\textwidth]{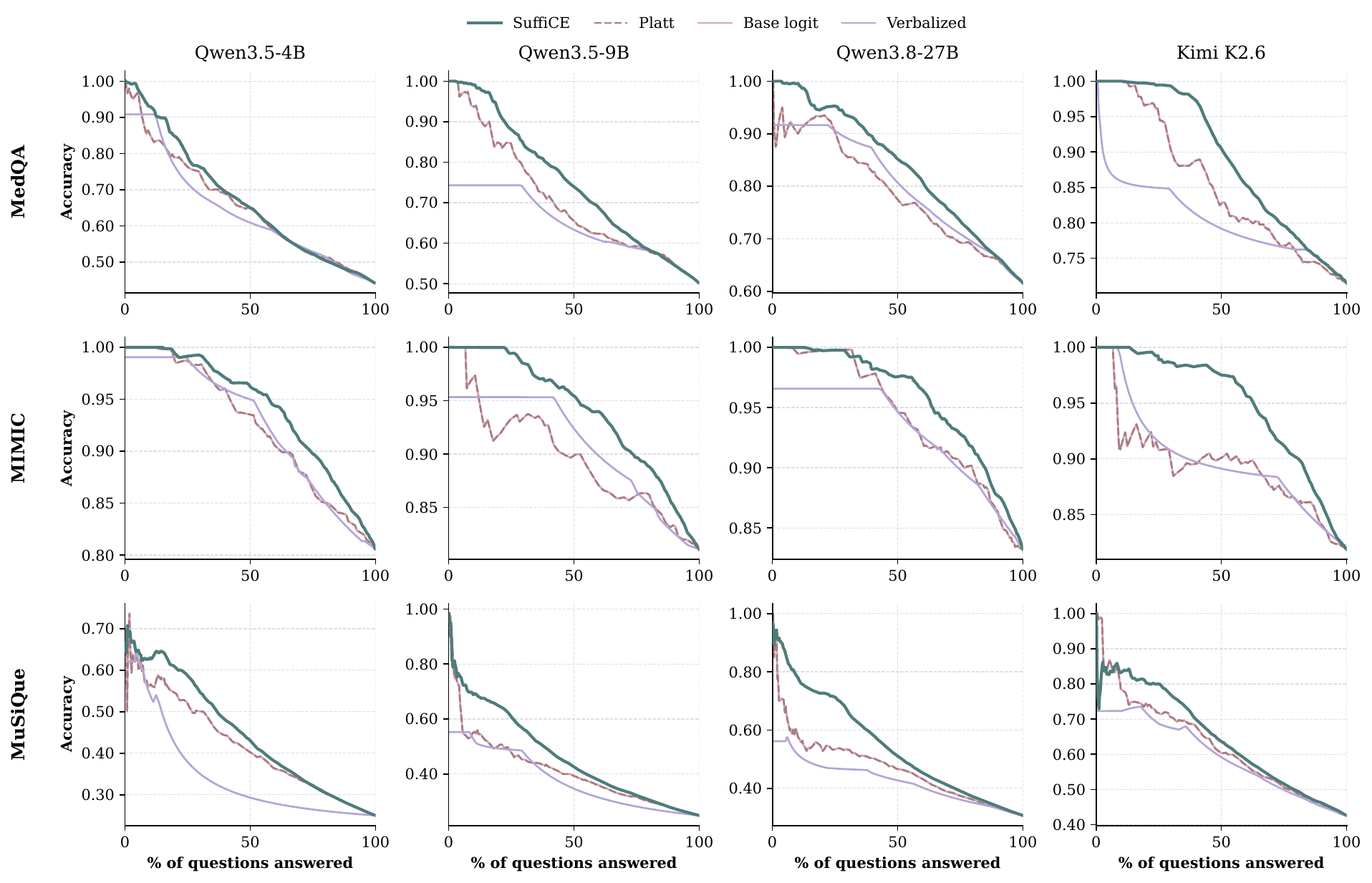}
  \caption{Selective answering on the test set of every dataset (rows) and model (columns). Each curve plots the accuracy of the answered questions against the percentage answered, answering the questions with the highest confidence first. Higher is better. Platt scaling preserves the ranking of the base logit, so their curves coincide.}
  \label{fig:coverage_grid}
\end{figure*}

Across all four answering models and three datasets, \ours{} has the highest accuracy at most percentages of questions answered, with the smallest margin for Qwen3.5-4B on the clinical datasets (Figure~\ref{fig:coverage_grid}, extending Table~\ref{tab:selective}).

\begin{table}[!ht]
  \centering
  \small
  \setlength{\tabcolsep}{2pt}
  \renewcommand{\arraystretch}{0.9}
  \renewcommand{\sd}[1]{\,{\tiny$\pm$#1}}
  \begin{tabularx}{\columnwidth}{@{}l*{3}{>{\centering\arraybackslash}X}@{}}
    \toprule
    Method & Brier $\downarrow$ & AUROC $\uparrow$ & $\rho_\Delta$ $\uparrow$ \\
    \midrule
    \multicolumn{4}{@{}l}{\textit{MIMIC}} \\
    Full only & 0.130\sd{0.005} & 0.745\sd{0.014} & 0.251\sd{0.024} \\
    \rowcolor{oursrow}\ours{} & \bestm{0.098}{0.010} & \bestm{0.819}{0.028} & \bestm{0.393}{0.034} \\
    \midrule
    \multicolumn{4}{@{}l}{\textit{MedQA}} \\
    Full only & 0.131\sd{0.004} & 0.782\sd{0.007} & 0.554\sd{0.008} \\
    \rowcolor{oursrow}\ours{} & \bestm{0.107}{0.005} & \bestm{0.815}{0.011} & \bestm{0.630}{0.012} \\
    \midrule
    \multicolumn{4}{@{}l}{\textit{MuSiQue}} \\
    Full only & 0.101\sd{0.006} & 0.794\sd{0.015} & 0.532\sd{0.049} \\
    \rowcolor{oursrow}\ours{} & \bestm{0.078}{0.002} & \bestm{0.828}{0.006} & \bestm{0.666}{0.015} \\
    \bottomrule
  \end{tabularx}
  \caption{Step-matched ablation for Qwen3.5-9B. The full-only estimator is trained for the same number of steps as \ours{}.}
  \label{tab:ablation_steps}
\end{table}

\begin{table*}[t]
  \centering
  \small
  \setlength{\tabcolsep}{3pt}
  \renewcommand{\sd}[1]{\,{\tiny$\pm$#1}}
  \begin{tabularx}{\textwidth}{@{}l*{6}{>{\centering\arraybackslash}X}@{}}
    \toprule
    & \multicolumn{2}{c}{\textbf{\textit{MIMIC}}} & \multicolumn{2}{c}{\textbf{\textit{MedQA}}} & \multicolumn{2}{c}{\textbf{\textit{MuSiQue}}} \\
    \cmidrule(lr){2-3} \cmidrule(lr){4-5} \cmidrule(l){6-7}
    Method & $\Delta$MAE $\downarrow$ & $\rho_{\text{within}}$ $\uparrow$ & $\Delta$MAE $\downarrow$ & $\rho_{\text{within}}$ $\uparrow$ & $\Delta$MAE $\downarrow$ & $\rho_{\text{within}}$ $\uparrow$ \\
    \midrule
    \multicolumn{7}{@{}l}{\textit{Kimi K2.6}} \\
    Base logit & 0.333\sd{0.000} & 0.217\sd{0.000} & 0.282\sd{0.000} & 0.290\sd{0.000} & 0.355\sd{0.000} & 0.164\sd{0.000} \\
    Verbalized & \bestm{0.182}{0.001} & 0.222\sd{0.007} & 0.193\sd{0.002} & 0.247\sd{0.004} & 0.367\sd{0.003} & 0.123\sd{0.029} \\
    Platt & 0.190\sd{0.000} & 0.210\sd{0.000} & 0.202\sd{0.000} & 0.290\sd{0.000} & 0.351\sd{0.000} & 0.173\sd{0.000} \\
    \rowcolor{oursrow}\ours{} & 0.188\sd{0.008} & \bestm{0.496}{0.022} & \bestm{0.188}{0.013} & \bestm{0.430}{0.059} & \bestm{0.292}{0.003} & \bestm{0.216}{0.021} \\
    \midrule
    \multicolumn{7}{@{}l}{\textit{Qwen3.8-27B}} \\
    Base logit & 0.260\sd{0.000} & 0.299\sd{0.000} & 0.281\sd{0.000} & 0.440\sd{0.000} & 0.353\sd{0.000} & 0.026\sd{0.000} \\
    Verbalized & 0.193\sd{0.000} & 0.248\sd{0.003} & 0.238\sd{0.001} & 0.480\sd{0.003} & 0.400\sd{0.003} & 0.138\sd{0.007} \\
    Platt & 0.198\sd{0.000} & 0.253\sd{0.000} & 0.219\sd{0.000} & 0.450\sd{0.000} & 0.337\sd{0.000} & 0.110\sd{0.000} \\
    \rowcolor{oursrow}\ours{} & \bestm{0.180}{0.006} & \bestm{0.497}{0.007} & \bestm{0.218}{0.009} & \bestm{0.634}{0.023} & \bestm{0.240}{0.010} & \bestm{0.257}{0.025} \\
    \midrule
    \multicolumn{7}{@{}l}{\textit{Qwen3.5-9B}} \\
    Base logit & 0.217\sd{0.000} & 0.137\sd{0.000} & 0.259\sd{0.000} & 0.398\sd{0.000} & 0.383\sd{0.000} & 0.190\sd{0.000} \\
    Verbalized & 0.251\sd{0.003} & 0.235\sd{0.016} & 0.256\sd{0.001} & 0.381\sd{0.004} & 0.343\sd{0.002} & 0.193\sd{0.004} \\
    Platt & 0.202\sd{0.000} & 0.204\sd{0.000} & 0.234\sd{0.000} & 0.433\sd{0.000} & 0.307\sd{0.000} & 0.195\sd{0.000} \\
    \rowcolor{oursrow}\ours{} & \bestm{0.182}{0.004} & \bestm{0.454}{0.107} & \bestm{0.205}{0.005} & \bestm{0.592}{0.024} & \bestm{0.243}{0.006} & \bestm{0.229}{0.015} \\
    \midrule
    \multicolumn{7}{@{}l}{\textit{Qwen3.5-4B}} \\
    Base logit & 0.269\sd{0.000} & 0.386\sd{0.000} & 0.264\sd{0.000} & 0.514\sd{0.000} & 0.309\sd{0.000} & 0.168\sd{0.000} \\
    Verbalized & 0.326\sd{0.003} & 0.383\sd{0.004} & 0.304\sd{0.007} & 0.409\sd{0.019} & 0.336\sd{0.001} & 0.123\sd{0.003} \\
    Platt & 0.201\sd{0.000} & 0.368\sd{0.000} & 0.236\sd{0.000} & 0.512\sd{0.000} & 0.304\sd{0.000} & 0.207\sd{0.000} \\
    \rowcolor{oursrow}\ours{} & \bestm{0.178}{0.014} & \bestm{0.487}{0.034} & \bestm{0.222}{0.010} & \bestm{0.530}{0.036} & \bestm{0.258}{0.004} & \bestm{0.230}{0.011} \\
    \bottomrule
  \end{tabularx}
  \caption{Evidence-state analysis for every answering model. \(\Delta\)MAE compares changes in confidence with changes in empirical success between states of the same problem, and \(\rho_{\text{within}}\) correlates confidence with success among problems that share a state type. The best result for each model is in bold.}
  \label{tab:info_states}
\end{table*}

\subsection{Step-Matched Ablation}
\label{app:ablation_steps}

The ablation of \S\ref{sec:ablation} trains the full-only estimator for the same number of epochs as \ours{}. Since it has one state per problem instead of four to eight, it then takes far fewer gradient steps. Here we instead train it for the same number of steps as \ours{}, with all other hyperparameters unchanged (Table~\ref{tab:ablation_steps}). \ours{} remains better on every metric and dataset, so the gap comes from training across evidence states, not from the number of updates.

\begin{table}[!ht]
  \centering
  \small
  \setlength{\tabcolsep}{0pt}
  \begin{tabular*}{\columnwidth}{@{}l@{\extracolsep{\fill}}ccc@{}}
    \toprule
    Metric & Base logit & Cross-domain & In-domain \\
    \midrule
    \multicolumn{4}{@{}l}{Train MIMIC $\rightarrow$ test MedQA} \\
    \quad Brier $\downarrow$ & 0.268\sd{0.000} & \textit{0.189}\sd{0.027} & \textbf{0.107}\sd{0.005} \\
    \quad AUROC $\uparrow$ & 0.746\sd{0.000} & \textit{0.779}\sd{0.009} & \textbf{0.815}\sd{0.011} \\
    \quad $\rho_\Delta$ $\uparrow$ & 0.367\sd{0.001} & \textit{0.401}\sd{0.027} & \textbf{0.630}\sd{0.012} \\
    \quad $\rho_{\text{within}}$ $\uparrow$ & 0.398\sd{0.000} & \textit{0.491}\sd{0.029} & \textbf{0.592}\sd{0.024} \\
    \midrule
    \multicolumn{4}{@{}l}{Train MedQA $\rightarrow$ test MIMIC} \\
    \quad Brier $\downarrow$ & \textit{0.124}\sd{0.000} & 0.167\sd{0.004} & \textbf{0.098}\sd{0.010} \\
    \quad AUROC $\uparrow$ & 0.689\sd{0.000} & \textit{0.746}\sd{0.016} & \textbf{0.819}\sd{0.028} \\
    \quad $\rho_\Delta$ $\uparrow$ & 0.247\sd{0.000} & \textit{0.316}\sd{0.005} & \textbf{0.393}\sd{0.034} \\
    \quad $\rho_{\text{within}}$ $\uparrow$ & 0.137\sd{0.000} & \textit{0.321}\sd{0.038} & \textbf{0.454}\sd{0.107} \\
    \bottomrule
  \end{tabular*}
  \caption{Cross-domain transfer between the clinical datasets (Qwen3.5-9B). Cross-domain and in-domain are \ours{} trained on the other dataset and on the test dataset itself. The best result in each row is in bold and the second best in italics.}
  \label{tab:transfer}
\end{table}

\subsection{Additional Evidence-State Analysis}
\label{app:info_state_analysis}

\paragraph{Ordering of pairs in \(\rho_\Delta\).} Each pair of states enters once, ordered from less to more evidence. The correlation subtracts the average change, so confidence that rises equally whenever evidence is added gets no credit, and \(\rho_\Delta\) rewards confidence that rises more for the evidence that raises success more.

We report two measures of evidence sensitivity beyond \(\rho_\Delta\) for every answering model (Table~\ref{tab:info_states}).
\begin{itemize}
  \item \textbf{\(\Delta\)MAE} (lower is better) tests whether confidence changes by the right amount as the evidence changes. For every pair of distinct states \(s,t\) of the same problem \(i\), the same pairs as \(\rho_\Delta\), it takes the absolute gap between the change in confidence and the change in empirical success, \(\bigl|(c_{i,t}-c_{i,s})-(\hat p_{i,t}-\hat p_{i,s})\bigr|\), and averages over all pairs in the test set. It is zero when confidence rises and falls by exactly as much as success. Where \(\rho_\Delta\) tests the direction of each change, \(\Delta\)MAE tests its size.
  \item \textbf{\(\rho_{\text{within}}\)} (higher is better) tests whether confidence separates more and less answerable problems that have the same kind of evidence. It groups the test set by state type, such as all \textsc{History} states (on MuSiQue, by question type and state), computes the Pearson correlation between confidence and empirical success within each group, and averages over groups. Every state in a group holds the same kind of evidence, so recognizing the state type alone does not raise it.
\end{itemize}

\ours{} has the lowest \(\Delta\)MAE in 11 of the 12 combinations of model and dataset, so its confidence changes by about as much as success does, not only in the same direction. The one exception is Kimi K2.6 on MIMIC, where verbalized confidence is slightly lower. For Kimi K2.6, success on MIMIC changes by at most 0.1 for about three quarters of the pairs of states, so a confidence that barely moves has a low \(\Delta\)MAE, and verbalized confidence moves about half as much as \ours{}. On \(\rho_\Delta\), which gives no credit for staying still, \ours{} is far better (Table~\ref{tab:main_results}). \ours{} also has the highest \(\rho_{\text{within}}\) for every model and dataset, with the largest margin on MIMIC, where whether the evidence suffices depends on the particular patient, so it does not merely learn an average confidence for each state type.

\subsection{Cross-Domain Transfer}
\label{app:transfer}

To test whether what \ours{} learns is specific to one dataset, we train it on one clinical dataset and evaluate it on the other without further training (Table~\ref{tab:transfer}). MIMIC asks for one of four diseases and MedQA for a free-text diagnosis, and success is far more common on MIMIC.

\paragraph{Which evidence matters transfers.} In both directions, the transferred estimator ranks states better than the base logit in AUROC and \(\rho_\Delta\). It also better separates more and less answerable cases that share a state type, recovering about half of the in-domain gain in \(\rho_{\text{within}}\) (\S\ref{app:info_state_analysis}) over the base logit in both directions. What \ours{} learns about which evidence settles a case thus carries over across datasets.

\paragraph{The overall success level does not transfer.} The transferred confidence stays close to the success rate of the training dataset. Trained on MIMIC, where most states succeed, it remains too confident on MedQA, though less so than the base logit. Trained on MedQA, it is too cautious on MIMIC and is worse calibrated than the base logit. This error shifts the level of confidence, not the order of states, so ranking transfers while calibration does not.

\clearpage
\raggedbottom
\section{Proofs}
\label{app:theory}

Proposition~\ref{prop:decomp} says that the error of a confidence splits into two parts: how far off it is on average for each problem, and how far its changes between states of the same problem are from the changes in success. Calibration constrains only the first part. Corollary~\ref{cor:bound} shows that the training loss bounds both parts. We first define the notation shared by both results, then restate and prove Proposition~\ref{prop:decomp} (\S\ref{app:proof_prop}) and Corollary~\ref{cor:bound} (\S\ref{app:proof_cor}).

\paragraph{Setup.} Let there be $n$ problems, indexed by $i=1,\ldots,n$. Problem $i$ consists of a question $q_i$, a pool of evidence $E_i$, and a gold answer $y^*_i$ (\S\ref{sec:formulation}). It has $m_i\geq 1$ evidence states, indexed by $s=1,\ldots,m_i$. State $s$ pairs the question with a subset $I_{i,s}\subseteq E_i$ of the evidence, $x_{i,s}=(q_i,I_{i,s})$. Let $N=\sum_{i=1}^n m_i$ be the total number of states. For a state $x$, let $p^*(x)\in[0,1]$ be the success probability of the answering policy $\pi_0$ (Eq.~\eqref{eq:pstar}) and $c_\theta(x)\in[0,1]$ the confidence of the estimator. 

Within problem $i$, we abbreviate $c_s=c_\theta(x_{i,s})$ and $p_s=p^*(x_{i,s})$, and define the error at state $s$ as $\epsilon_s=c_s-p_s$. The problem averages of the confidence, the success probability, and the error are
\begin{gather*}
    \bar c_{\theta,i}=\frac{1}{m_i}\sum_{s=1}^{m_i}c_s,\qquad
    \bar p^*_i=\frac{1}{m_i}\sum_{s=1}^{m_i}p_s,\\
    \bar\epsilon=\bar c_{\theta,i}-\bar p^*_i .
\end{gather*}
For any $s,t\in\{1,\ldots,m_i\}$, the change gap
\begin{equation*}
    \delta_{st}=(c_t-c_s)-(p_t-p_s)
\end{equation*}
is the difference between the change in confidence and the change in success probability from state $s$ to state $t$. Finally, $\mathbb{E}$ denotes the uniform average over all $N$ states, and $\mathbb{E}_i$ the average of a per-problem quantity over problems, with problem $i$ weighted by $m_i/N$, its share of all states. With this weight, averaging any function $f$ of a state over all states equals first averaging it within each problem and then applying $\mathbb{E}_i$,
\begin{equation}
\begin{split}
    \mathbb{E}[f]
    &= \frac{1}{N}\sum_{i=1}^{n}\sum_{s=1}^{m_i}f(x_{i,s}) \\
    &= \sum_{i=1}^{n}\frac{m_i}{N}\cdot\frac{1}{m_i}\sum_{s=1}^{m_i}f(x_{i,s}) \\
    &= \sum_{i=1}^{n}\frac{m_i}{N}\,\bar f_i
    = \mathbb{E}_i\bigl[\bar f_i\bigr],
\end{split}
\label{eq:group}
\end{equation}
where $\bar f_i=\frac{1}{m_i}\sum_{s}f(x_{i,s})$ is the average of $f$ over the states of problem $i$. A plain average over problems, $\frac{1}{n}\sum_i\bar f_i$, agrees with $\mathbb{E}[f]$ only when all problems have the same number of states.

\subsection{Proof of Proposition~\ref{prop:decomp}}
\label{app:proof_prop}

\paragraph{Proposition~\ref{prop:decomp} (restated).} \textit{For any confidence $c_\theta$,}
\begin{equation*}
\begin{split}
    \mathbb{E}\bigl[(c_\theta-p^*)^2\bigr]
    &= \mathbb{E}_i\bigl[(\bar c_{\theta,i}-\bar p^*_i)^2\bigr] \\
    &\quad + \mathbb{E}_i\Bigl[\tfrac{1}{2m_i^2}\textstyle\sum_{s=1}^{m_i}\sum_{t=1}^{m_i}\delta_{st}^2\Bigr].
\end{split}
\end{equation*}
\textit{The left side is the mean squared error over all states. The first term on the right is the error across problems, in each problem's average success. The second is the error within each problem, half the average squared change gap over all ordered pairs of its states.}

\paragraph{Proof.} We prove the identity for a single problem $i$ and then average over problems. All sums over $s$ and $t$ run from $1$ to $m_i$.

\textit{Step 1: split the error into its average and its spread.} Writing $\epsilon_s=\bar\epsilon+(\epsilon_s-\bar\epsilon)$ and expanding the square,
\begin{equation*}
\begin{split}
    \frac{1}{m_i}\sum_{s}\epsilon_s^2
    &= \bar\epsilon^{\,2} + \frac{2\bar\epsilon}{m_i}\sum_{s}(\epsilon_s-\bar\epsilon) \\
    &\quad + \frac{1}{m_i}\sum_{s}(\epsilon_s-\bar\epsilon)^2 .
\end{split}
\end{equation*}
The middle term is zero. Since $\bar\epsilon=\bar c_{\theta,i}-\bar p^*_i=\frac{1}{m_i}\sum_s\epsilon_s$ is the average of the errors, the errors above the average exactly cancel those below it,
\begin{equation*}
    \sum_{s}(\epsilon_s-\bar\epsilon)=\sum_{s}\epsilon_s-m_i\bar\epsilon=m_i\bar\epsilon-m_i\bar\epsilon=0 .
\end{equation*}

\textit{Step 2: rewrite the spread with pairs of states.} The spread is the last term of Step~1, $\frac{1}{m_i}\sum_s(\epsilon_s-\bar\epsilon)^2$, the average squared deviation of the errors from their mean, that is, their variance over the states of problem $i$. It is zero exactly when every state of the problem has the same error. We now show that it equals half the average squared difference between the errors of two states. Let $d_s=\epsilon_s-\bar\epsilon$, so $\sum_s d_s=0$ by Step~1. Expanding $(d_t-d_s)^2=d_t^2+d_s^2-2d_sd_t$ and summing over all $s$ and $t$,
\begin{equation*}
\begin{split}
    \sum_{s,t}(\epsilon_t-\epsilon_s)^2
    &= \sum_{s,t}(d_t-d_s)^2 \\
    &= 2m_i\sum_{s}d_s^2 - 2\Bigl(\sum_{s}d_s\Bigr)^2 \\
    &= 2m_i\sum_{s}d_s^2 \\
    &= 2m_i\sum_{s}(\epsilon_s-\bar\epsilon)^2 .
\end{split}
\end{equation*}
Dividing by $2m_i^2$ gives $\frac{1}{m_i}\sum_s(\epsilon_s-\bar\epsilon)^2=\frac{1}{2m_i^2}\sum_{s,t}(\epsilon_t-\epsilon_s)^2$.

\textit{Step 3: identify the change gap.} Rearranging, $\epsilon_t-\epsilon_s=(c_t-p_t)-(c_s-p_s)=(c_t-c_s)-(p_t-p_s)=\delta_{st}$. Combining the three steps, for every problem $i$,
\begin{equation*}
\begin{split}
    \frac{1}{m_i}\sum_{s}\epsilon_s^2
    &\;\overset{\text{\tiny Step 1}}{=}\; \bar\epsilon^{\,2} + \frac{1}{m_i}\sum_{s}(\epsilon_s-\bar\epsilon)^2 \\
    &\;\overset{\text{\tiny Step 2}}{=}\; \bar\epsilon^{\,2} + \frac{1}{2m_i^2}\sum_{s,t}(\epsilon_t-\epsilon_s)^2 \\
    &\;\overset{\text{\tiny Step 3}}{=}\; (\bar c_{\theta,i}-\bar p^*_i)^2 + \frac{1}{2m_i^2}\sum_{s,t}\delta_{st}^2 ,
\end{split}
\end{equation*}
where the last line substitutes $\bar\epsilon=\bar c_{\theta,i}-\bar p^*_i$ and $\epsilon_t-\epsilon_s=\delta_{st}$. The left side is the mean squared error of problem $i$. On the right, the first term is the squared error of the problem's average confidence, and the second is half the average squared change gap over its ordered pairs of states.

\textit{Step 4: average over problems.} Steps 1--3 give the identity for one problem. The proposition averages over all states, so we group the states by problem and weight each problem by its share of states, using Eq.~\eqref{eq:group} with $f=(c_\theta-p^*)^2$. Substituting the per-problem identity for each problem's mean squared error and splitting the sum,
\begin{equation*}
\begin{split}
    &\mathbb{E}\bigl[(c_\theta-p^*)^2\bigr]
    = \sum_{i}\frac{m_i}{N}\Bigl[\frac{1}{m_i}\sum_{s}\epsilon_s^2\Bigr] \\
    &= \sum_{i}\frac{m_i}{N}\Bigl[(\bar c_{\theta,i}-\bar p^*_i)^2 + \frac{1}{2m_i^2}\sum_{s,t}\delta_{st}^2\Bigr] \\
    &= \sum_{i}\frac{m_i}{N}(\bar c_{\theta,i}-\bar p^*_i)^2
     + \sum_{i}\frac{m_i}{N}\cdot\frac{1}{2m_i^2}\sum_{s,t}\delta_{st}^2 \\
    &= \mathbb{E}_i\bigl[(\bar c_{\theta,i}-\bar p^*_i)^2\bigr]
     + \mathbb{E}_i\Bigl[\tfrac{1}{2m_i^2}\textstyle\sum_{s,t}\delta_{st}^2\Bigr],
\end{split}
\end{equation*}
where the last line uses the definition of $\mathbb{E}_i$ as the average over problems with weight $m_i/N$. This is Eq.~\eqref{eq:decomp}. \hfill$\square$

\paragraph{Remarks.} \textit{Calibration alone does not remove the second term.} Suppose $c_\theta$ gives every state of a problem the same value, for example the problem's average success $\bar p^*_i$. Then $c_t-c_s=0$, so $\delta_{st}=-(p_t-p_s)$, and the second term becomes $\frac{1}{2m_i^2}\sum_{s,t}(p_t-p_s)^2$, which by the identity of Step~2, applied to $p_s$ in place of $\epsilon_s$, equals the variance of $p^*$ over the states of problem $i$. This value does not depend on which constant is chosen, so a confidence that ignores the evidence cannot reduce it, even if it is perfectly calibrated.

\textit{The Brier score measures the same error.} The Brier score (\S\ref{sec:metrics}) compares the confidence with the empirical success rate $\hat p(x)$ rather than with $p^*(x)$. Since $\hat p(x)$ averages $K$ independent grades, each equal to $1$ with probability $p^*(x)$, it satisfies $\mathbb{E}[\hat p(x)\mid x]=p^*(x)$ and $\mathrm{Var}(\hat p(x)\mid x)=p^*(x)(1-p^*(x))/K$. For a confidence computed from $x$ alone, the cross term vanishes, and
\begin{equation*}
\begin{split}
    \mathbb{E}\bigl[(c_\theta-\hat p)^2\bigr]
    &= \mathbb{E}\bigl[(c_\theta-p^*)^2\bigr] \\
    &\quad + \tfrac{1}{K}\,\mathbb{E}\bigl[p^*(1-p^*)\bigr],
\end{split}
\end{equation*}
where the expectation on the left also averages over the sampled grades. The last term does not depend on the confidence, so differences in Brier score between methods reflect differences in the error that Proposition~\ref{prop:decomp} decomposes.

\subsection{Proof of Corollary~\ref{cor:bound}}
\label{app:proof_cor}

\paragraph{Corollary~\ref{cor:bound} (restated).} \textit{For any confidence $c_\theta$, let $\mathbb{E}[\mathcal L]$ be the loss of Eq.~\eqref{eq:loss} averaged over all evidence states and over the $K$ sampled answers that determine $\hat p(x)$, and $\mathcal L^*$ the minimum expected loss over all confidences, reached exactly at $c_\theta=p^*$. Then}
\begin{equation*}
    \mathbb{E}\bigl[(c_\theta-p^*)^2\bigr]\leq\tfrac12\bigl(\mathbb{E}[\mathcal L]-\mathcal L^*\bigr),
\end{equation*}
\textit{so each term of Proposition~\ref{prop:decomp} is at most $\frac12(\mathbb{E}[\mathcal L]-\mathcal L^*)$.}

\paragraph{Proof.} Recall the soft-label binary cross-entropy of Eq.~\eqref{eq:loss}, at each evidence state $x$,
\begin{equation*}
\begin{split}
    \mathcal L(x)
    ={}&
    -\hat p(x)\log c_\theta(x) \\
    &-
    \bigl(1-\hat p(x)\bigr)
    \log\bigl(1-c_\theta(x)\bigr).
\end{split}
\end{equation*}
Let $H(p)=-p\log p-(1-p)\log(1-p)$ be the binary entropy and $\mathrm{KL}(p\,\|\,c)=p\log\frac{p}{c}+(1-p)\log\frac{1-p}{1-c}$ the divergence between Bernoulli distributions with means $p$ and $c$, with the convention $0\log 0=0$. The confidence of \ours{} is a softmax and so lies strictly between 0 and 1, which keeps every logarithm finite. The loss is linear in $\hat p(x)$, so taking the expectation over the sampled grades replaces $\hat p(x)$ by $\mathbb{E}[\hat p(x)\mid x]=p^*(x)$. At each state, writing $p^*$ and $c_\theta$ for $p^*(x)$ and $c_\theta(x)$, the expected loss is therefore
\begin{equation*}
    \mathbb{E}\bigl[\mathcal L(x)\mid x\bigr]
    = -p^*\log c_\theta-(1-p^*)\log(1-c_\theta).
\end{equation*}
To separate the part that depends on $c_\theta$ from the part that does not, we use the entropy of the target, $H(p^*)=-p^*\log p^*-(1-p^*)\log(1-p^*)$. By this definition, $H(p^*)+p^*\log p^*+(1-p^*)\log(1-p^*)=0$, so adding these terms to the expected loss leaves it unchanged:
\begin{equation*}
\begin{split}
    \mathbb{E}\bigl[\mathcal L(x)\mid x\bigr]
    &= H(p^*)
    + p^*\log p^* \\
    &\quad +(1-p^*)\log(1-p^*) \\
    &\quad - p^*\log c_\theta \\
    &\quad -(1-p^*)\log(1-c_\theta).
\end{split}
\end{equation*}
Grouping the two terms with the factor $p^*$ and the two with the factor $1-p^*$,
\begin{equation*}
\begin{split}
    \mathbb{E}\bigl[\mathcal L(x)\mid x\bigr]
    &= H(p^*)
    + p^*\bigl(\log p^*-\log c_\theta\bigr) \\
    &\quad + (1-p^*)\bigl(\log(1-p^*) \\
    &\qquad -\log(1-c_\theta)\bigr),
\end{split}
\end{equation*}
and writing each difference of logarithms as the logarithm of a ratio,
\begin{equation*}
\begin{split}
    \mathbb{E}\bigl[\mathcal L(x)\mid x\bigr]
    &= H(p^*)
    + p^*\log\frac{p^*}{c_\theta} \\
    &\quad + (1-p^*)\log\frac{1-p^*}{1-c_\theta} \\
    &= H(p^*)+\mathrm{KL}(p^*\,\|\,c_\theta).
\end{split}
\end{equation*}
Averaging this identity over all evidence states gives
\begin{equation*}
    \mathbb{E}[\mathcal L]=\mathbb{E}\bigl[H(p^*)\bigr]+\mathbb{E}\bigl[\mathrm{KL}(p^*\,\|\,c_\theta)\bigr].
\end{equation*}
The first term depends only on the answering policy, not on the confidence, so no choice of $c_\theta$ can change it. The second term is an average of divergences, each nonnegative and zero only when $c_\theta(x)=p^*(x)$. Every confidence therefore has $\mathbb{E}[\mathcal L]\geq\mathbb{E}[H(p^*)]$, with equality exactly when $c_\theta(x)=p^*(x)$ at every state. Hence the minimum expected loss is $\mathcal L^*=\mathbb{E}[H(p^*)]$, reached exactly at $c_\theta=p^*$, and
\begin{equation*}
    \mathbb{E}[\mathcal L]-\mathcal L^*=\mathbb{E}\bigl[\mathrm{KL}(p^*\,\|\,c_\theta)\bigr].
\end{equation*}
Pinsker's inequality for Bernoulli distributions states $\mathrm{KL}(p\,\|\,c)\geq 2(p-c)^2$. Applying it at every state and averaging,
\begin{equation*}
\begin{split}
    \mathbb{E}[\mathcal L]-\mathcal L^*
    &= \mathbb{E}\bigl[\mathrm{KL}(p^*\,\|\,c_\theta)\bigr] \\
    &\geq 2\,\mathbb{E}\bigl[(c_\theta-p^*)^2\bigr],
\end{split}
\end{equation*}
which is Eq.~\eqref{eq:bound}. By Proposition~\ref{prop:decomp}, the two terms are nonnegative and sum to $\mathbb{E}[(c_\theta-p^*)^2]$, so each is at most $\frac12(\mathbb{E}[\mathcal L]-\mathcal L^*)$. \hfill$\square$

For \ours{}, this means that lowering the training loss lowers an upper bound on both terms of Proposition~\ref{prop:decomp}, the error in each problem's average success and the error in how confidence changes between states of the same problem. The bound holds only over the states the loss is averaged over, which is why \ours{} trains on many evidence states of each problem rather than on full states alone (\S\ref{sec:ablation}).

\flushbottom
\twocolumn[{%
  \centering
  \small
  \setlength{\tabcolsep}{4pt}
  \begin{tabularx}{\textwidth}{@{}l*{5}{>{\centering\arraybackslash}X}@{}}
    \toprule
    Method & Pre-answer & Trained & Controlled evidence subsets & Within-problem evaluation & Own token probability \\
    \midrule
    P(IK)~\citep{kadavath2022language} & \yesmark & \yesmark & \nomark & \nomark & \nomark \\
    Pre-generation probes~\citep{lugoloobi2026encode} & \yesmark & \yesmark & \nomark & \nomark & \nomark \\
    Calibration tuning~\citep{kapoor2024LLM} & \nomark & \yesmark & \nomark & \nomark & \yesmark \\
    Rewarding Doubt~\citep{bani2026rewarding} & \nomark & \yesmark & \nomark & \nomark & \nomark \\
    LA-CDM~\citep{bani2026lacdm} & \nomark & \yesmark & \nomark & \nomark & \nomark \\
    P(Sufficient)~\citep{zhang-etal-2026-confidence-estimation} & \nomark & \nomark & \nomark & \nomark & \yesmark \\
    \midrule
    \ours{} & \yesmark & \yesmark & \yesmark & \yesmark & \yesmark \\
    \bottomrule
  \end{tabularx}
  \captionof{table}{Comparison with the closest prior work along the five properties defined below. \yesmark{} yes, \nomark{} no.}
  \label{tab:related_comparison}
  \vspace{1.2em}
}]
\section{Comparison with Related Work}
\label{app:related_comparison}

We compare our setting with the closest prior work on confidence estimation along five properties (Table~\ref{tab:related_comparison}).
\begin{itemize}
  \item \textbf{Pre-answer}: the confidence is produced before any answer is generated, so it can decide whether to answer or to gather more evidence.
  \item \textbf{Trained}: the confidence estimator is trained, rather than read from an untrained model or elicited by prompting.
  \item \textbf{Controlled evidence subsets}: the same problem is observed with different combinations of evidence, not only with evidence that accumulates in a fixed order.
  \item \textbf{Within-problem evaluation}: the evaluation tests whether changes in confidence follow changes in success across states of the same problem.
  \item \textbf{Own token probability}: the confidence is the model's next-token probability, with no added classifier head or verbalized number.
\end{itemize}
P(IK)~\citep{kadavath2022language} trains an added value head to predict, from the question alone, whether the model will answer correctly. It adds source documents and hints only at evaluation and does not relate the change in P(IK) to the change in accuracy. Pre-generation probes~\citep{lugoloobi2026encode} train linear probes on the model's activations before generation to predict success, with each problem shown in a single fixed prompt. Calibration tuning~\citep{kapoor2024LLM} fine-tunes the model to judge whether a proposed answer is correct, read from its token probability. Rewarding Doubt~\citep{bani2026rewarding} uses reinforcement learning to train the model to state a confidence from 0 to 10 in an answer it has already generated, with each question asked without context. LA-CDM~\citep{bani2026lacdm} trains a stated confidence in a diagnosis the agent has already proposed, and the evidence is limited to the tests its agent chose to request. P(Sufficient)~\citep{zhang-etal-2026-confidence-estimation} prompts the model for the probability that the information so far entails its answer. Its clues arrive in a fixed order, and its evaluation measures per-turn calibration and whether confidence rises over turns.

Each prior method shares some of these properties, and ours is the only one that has all five. No prior method trains the model's own token probability before an answer exists. P(IK)~\citep{kadavath2022language} and pre-generation probes~\citep{lugoloobi2026encode} estimate success before answering but read it from an added head or probe, while calibration tuning~\citep{kapoor2024LLM} and P(Sufficient)~\citep{zhang-etal-2026-confidence-estimation} use the model's token probability but score an answer it has already given. \ours{} instead trains the model's own probability of choosing to answer, so the same next-token probability that would decide whether to answer or gather more evidence is calibrated, with no extra head and no answer to generate first. We choose this formulation because the model's next-token distribution already assigns a probability to answering versus seeking more evidence, so no classifier head is added and the confidence is the model's own preference between the two actions. A full-only estimator with the same formulation falls short of \ours{} (\S\ref{sec:ablation}), so the gains come from training across evidence states. Two further properties set \ours{} apart, and no prior method has either. Training on controlled evidence subsets of the same problem lets the estimator learn which evidence decides success, rather than only which problems are hard (\S\ref{sec:ablation}). Evaluating within each problem tests this directly, showing that confidence rises with the decisive evidence and not with every piece added (\S\ref{sec:info_state_analysis}).

\clearpage
\section{Case Studies}
\label{app:case_studies}

\subsection{MuSiQue Case Studies}
\label{app:case_study}

We examine two branching (3hop2) questions from the MuSiQue test set in detail (Tables~\ref{tab:case_study} and~\ref{tab:case_study_lotr}). In both, the branches \(e_1\) and \(e_2\) answer two independent sub-questions, and only the final hop \(e_3\) combines their answers into the requested date.

\paragraph{Case 1: When the Republicans took control of the Senate.} The branches establish that Alaska supports the Republicans (\(e_1\)) and that the Senate sets its own rules (\(e_2\)), but only \(e_3\) gives the date on which the Republicans regained the Senate. With both branches shown, no sampled answer is correct. The base logit nonetheless jumps from at most 0.02 to 0.50, treating the two branches as progress despite the lack of sampled success. \ours{} stays at or below 0.02, matching this low success rate. Once \(e_3\) is shown, 20\% of sampled answers are correct. \ours{} rises to 0.19, closely matching this rate, while the base logit falls to 0.11 and becomes underconfident.

\paragraph{Case 2: When the Dutch reached New Zealand.} The branches establish that The Lord of the Rings was filmed in New Zealand (\(e_1\)) and that the Dutch first brought enslaved people to New York (\(e_2\)), but only \(e_3\) gives the date on which the Dutch navigator Abel Tasman reached New Zealand. Again no sampled answer is correct with both branches, yet the base logit climbs with each branch to 0.98. \ours{} reaches only 0.16. With full evidence, 30\% of sampled answers are correct. \ours{} gives 0.34, again close to this rate, while the base logit stays at 0.96 and is far overconfident.

\paragraph{Comparing the two full states.} The two questions share the same structure and both show all three supporting paragraphs, yet they differ in how often they are answered correctly. \ours{} follows this difference, matching each success rate within 0.04, whereas the base logit misses in opposite directions. The confidence of \ours{} is not fixed by the evidence being complete, so it judges not only whether the reasoning chain is complete but also how likely the answering model is to succeed once it is.

\begin{table}[!ht]
  \centering
  \small
  \renewcommand{\arraystretch}{1.1}
  \begin{tabularx}{\columnwidth}{@{}l@{\hspace{5pt}}>{\raggedright\arraybackslash}X@{}}
    \toprule
    \multicolumn{2}{@{}>{\raggedright\arraybackslash}p{\columnwidth}@{}}{\textit{When did the party Alaska generally supports take control of the determiner of rules of the US House and US Senate?} \textbf{Answer:} January 2015} \\
    \midrule
    \enode{1}{hoplight} & Alaska has voted Republican in nearly every presidential election since statehood. \mbox{\color{factgray}$\rightarrow$ Republicans} \\
    \enode{2}{hoplight} & Each chamber of Congress sets its own rules (Article I), and the Senate adopts its Standing Rules. \mbox{\color{factgray}$\rightarrow$ the Senate} \\
    \enode{3}{finallight} & Republicans regained the Senate majority in the Congress that began in January 2015, having last held it in January 2007. \mbox{\color{factgray}$\rightarrow$ January 2015} \\
    \bottomrule
  \end{tabularx}

  \vspace{6pt}
  \renewcommand{\arraystretch}{1.0}
  \setlength{\tabcolsep}{3pt}
  \begin{tabularx}{\columnwidth}{@{}l*{4}{>{\hsize=0.9\hsize\centering\arraybackslash}X}>{\hsize=1.4\hsize\centering\arraybackslash}X@{}}
    \toprule\addlinespace[6pt]
    Evidence & \enodenone & \enodesm{1}{hoplight} & \enodesm{2}{hoplight} & \hlnode{hlmqT}{\enodesm{1}{hoplight}{+}\enodesm{2}{hoplight}} & \enodesm{1}{hoplight}{+}\enodesm{2}{hoplight}{+}\enodesm{3}{finallight}  \\
    \midrule
    Success & 0.10 & 0.00 & 0.00 & 0.00 & 0.20  \\
    Base     & 0.02 & 0.02 & 0.00 & 0.50 & 0.11  \\
    \ours{}  & 0.01 & 0.01 & 0.01 & \hlnode{hlmqB}{0.02} & 0.19  \\
    \bottomrule
  \end{tabularx}
  \hlcolumn{hlmq}{17pt}%
  \caption{Case study on a 3hop2 MuSiQue question (Qwen3.5-9B), where the branches \(e_1\) and \(e_2\) together lead to the final hop \(e_3\). Columns are evidence states, marked by which supporting facts are shown (\enodenone{} for none).}
  \label{tab:case_study}
\end{table}

\begin{table}[t]
  \centering
  \small
  \renewcommand{\arraystretch}{1.0}
  \begin{tabularx}{\columnwidth}{@{}l@{\hspace{5pt}}>{\raggedright\arraybackslash}X@{}}
    \toprule
    \multicolumn{2}{@{}>{\raggedright\arraybackslash}p{\columnwidth}@{}}{\textit{When did the people who first imported slaves to New York come to the country where The Lord of the Rings was filmed?} \textbf{Answer:} 13 December 1642} \\
    \midrule
    \enode{1}{hoplight} & The Lord of the Rings trilogy was filmed entirely in New Zealand. \mbox{\color{factgray}$\rightarrow$ New Zealand} \\
    \enode{2}{hoplight} & Enslaved Africans were first brought into New York by the Dutch, and later the English. \mbox{\color{factgray}$\rightarrow$ the Dutch} \\
    \enode{3}{finallight} & Dutch navigator Abel Tasman was the first European to sight New Zealand, on 13 December 1642. \mbox{\color{factgray}$\rightarrow$ 13 December 1642} \\
    \bottomrule
  \end{tabularx}

  \vspace{3pt}
  \renewcommand{\arraystretch}{1.0}
  \setlength{\tabcolsep}{2.5pt}
  \begin{tabularx}{\columnwidth}{@{}l*{4}{>{\hsize=0.9\hsize\centering\arraybackslash}X}>{\hsize=1.4\hsize\centering\arraybackslash}X@{}}
    \toprule\addlinespace[4pt]
    Evidence & \enodenone & \enodesm{1}{hoplight} & \enodesm{2}{hoplight} & \hlnode{hlapT}{\enodesm{1}{hoplight}{+}\enodesm{2}{hoplight}} & \enodesm{1}{hoplight}{+}\enodesm{2}{hoplight}{+}\enodesm{3}{finallight}  \\
    \midrule
    Success & 0.00 & 0.00 & 0.00 & 0.00 & 0.30  \\
    Base     & 0.12 & 0.15 & 0.65 & 0.98 & 0.96  \\
    \ours{}  & 0.02 & 0.03 & 0.06 & \hlnode{hlapB}{0.16} & 0.34  \\
    \bottomrule
  \end{tabularx}
  \hlcolumn{hlap}{17pt}%
  \caption{A second case study on a 3hop2 MuSiQue question (Qwen3.5-9B), in the format of Table~\ref{tab:case_study}. The shaded state has both branches but not the final hop.}
  \label{tab:case_study_lotr}
\end{table}

\subsection{Full Evidence States of the Case Studies}
\label{app:examples}

We give the complete inputs behind the clinical case study of \S\ref{sec:info_state_analysis} and the first MuSiQue case study of \S\ref{app:case_study} word for word, with the evidence states of each example and the evidence they contain (Tables~\ref{tab:full_clinical} and~\ref{tab:full_musique}). The state names and evidence labels, such as \(\{1,4\}\) or \(e_1\), are defined only for analysis and are never shown to the model. The model sees only the question and the evidence as it appears in the input, with the section headings of the clinical case but no indication of which state it is in.

\begin{table*}[t]
  \centering
  \footnotesize
  \renewcommand{\arraystretch}{1.1}
  \begin{tabularx}{\textwidth}{@{}l@{\hspace{6pt}}>{\raggedright\arraybackslash}X@{}}
    \toprule
    Objective & Assess and diagnose the infant brought in by a concerned mother due to bilious vomiting and inability to pass meconium, with observed abdominal distension. \textbf{Answer:} Hirschsprung's disease \\
    \midrule
    \cnum{1}{histstrong}~\textsc{History} & \textit{Demographics:} 2-day-old male infant. \textit{History:} The infant was born via a home birth without complications. The mother reports that the infant has had bilious vomiting and has not been able to pass meconium since birth. She also notes that the infant's abdomen appears distended. \textit{Primary symptom:} Bilious vomiting. \textit{Secondary symptoms:} Inability to pass meconium, Abdominal distension. \textit{Past medical history:} Newborn, no prior medical history. Birth was uneventful. \textit{Social history:} The infant is the first child of the family. No known family history of genetic or congenital diseases. \textit{Review of systems:} The infant shows signs of distress, possibly due to abdominal discomfort. \\
    \cnum{2}{examstrong}~\textsc{PE} & \textit{Vital signs:} Temperature 37.5$^\circ$C (99.5$^\circ$F), Blood pressure not applicable for age, Heart rate 145 bpm, Respiratory rate 40 breaths/min. \textit{Abdominal examination:} Inspection: Noticeable distension. Auscultation: Decreased bowel sounds. Percussion: Tympanic sound noted, indicating gas. Palpation: The abdomen is firm and distended, with palpable bowel loops; the infant shows signs of discomfort during palpation. \\
    \cnum{3}{nonimgstrong}~\textsc{Biopsy} & \textit{Rectal biopsy, histopathology:} Absence of ganglion cells in the submucosa and myenteric plexus, consistent with Hirschsprung's disease. \\
    \cnum{4}{imgstrong}~\textsc{Imaging} & \textit{Abdominal X-ray, findings:} Dilated loops of bowel with no gas in the rectum, suggestive of a distal obstruction. \\
    \bottomrule
  \end{tabularx}

  \vspace{6pt}
  \setlength{\tabcolsep}{6pt}
  \begin{tabular}{@{}ll@{}}
    \toprule
    State & Sections shown \\
    \midrule
    \(\{1\}\)       & \textsc{History} \\
    \(\{1,2\}\)     & \textsc{History}, \textsc{PE} \\
    \(\{1,4\}\)     & \textsc{History}, \textsc{Imaging} \\
    \(\{1,3,4\}\)   & \textsc{History}, \textsc{Biopsy}, \textsc{Imaging} \\
    \(\{1,2,3,4\}\) & \textsc{History}, \textsc{PE}, \textsc{Biopsy}, \textsc{Imaging} \\
    \bottomrule
  \end{tabular}
  \caption{Full input of the MedQA case of Table~\ref{tab:case_study_clinical} and the evidence states shown there. Each state shows the objective followed by the listed sections, verbatim.}
  \label{tab:full_clinical}
\end{table*}

\begin{table*}[h]
  \centering
  \small
  \renewcommand{\arraystretch}{1.1}
  \begin{tabularx}{\textwidth}{@{}l@{\hspace{6pt}}>{\raggedright\arraybackslash}X@{}}
    \toprule
    Question & When did the party Alaska generally supports take control of the determiner of rules of the US House and US Senate? \textbf{Answer:} January 2015 \\
    \midrule
    \enodesm{1}{hoplight} & \textbf{Alaska.} Alaska regularly supports Republicans in presidential elections and has done so since statehood. Republicans have won the state's electoral college votes in all but one election that it has participated in (1964). No state has voted for a Democratic presidential candidate fewer times. Alaska was carried by Democratic nominee Lyndon B. Johnson during his landslide election in 1964, while the 1960 and 1968 elections were close. Since 1972, however, Republicans have carried the state by large margins. In 2008, Republican John McCain defeated Democrat Barack Obama in Alaska, 59.49\% to 37.83\%. McCain's running mate was Sarah Palin, the state's governor and the first Alaskan on a major party ticket. Obama lost Alaska again in 2012, but he captured 40\% of the state's vote in that election, making him the first Democrat to do so since 1968. \\
    \enodesm{2}{hoplight} & \textbf{Standing Rules of the United States Senate.} The Standing Rules of the Senate are the parliamentary procedures adopted by the United States Senate that govern its procedure. The Senate's power to establish rules derives from Article One, Section 5 of the United States Constitution: ``Each House may determine the rules of its proceedings\ldots'' \\
    \enodesm{3}{finallight} & \textbf{2014 United States Senate elections.} The Republicans regained the majority of the Senate in the 114th Congress, which started in January 2015; the Republicans had not controlled the Senate since January 2007. They had needed a net gain of at least six seats to obtain a majority. They held all of their seats, and gained nine Democratic - held seats. Republicans defeated five Democratic incumbents: \\
    \dnodesm{1} & \textbf{Democratic Party of Oregon.} The Democratic Party of Oregon, based in Portland, is the official Oregon affiliate of the United States Democratic Party. It is recognized by the state of Oregon as a major political party, along with the Oregon Republican Party. The State Central Committee, made up of two delegates elected from each of Oregon's 36 counties and one additional delegate for every 15,000 registered Democrats, is the main authoritative body of the party. After Oregon was admitted into the Union in 1859, the Democratic party controlled the state. Oregon elected twice as many Democrats as Republicans between 1859 and 1879 in statewide elections for governor, secretary of state, state treasurer, and congressmen. The party holds 38 members in the State House that has 60 representatives total, and 18 members in the State Senate, out of 30 delegates total. The party also holds the Governor's office (Kate Brown), Attorney General (Ellen Rosenblum), Labor Commissioner (Val Hoyle), and State Treasurer (Tobias Read). The Democrats also have both U.S senate positions in their state and send four of the five U.S House representatives from Oregon to D.C. \\
    \dnodesm{2} & \textbf{Tammany Hall.} Tammany Hall, also known as the Society of St. Tammany, the Sons of St. Tammany, or the Columbian Order, was a New York City political organization founded in 1786 and incorporated on May 12, 1789, as the Tammany Society. It was the Democratic Party political machine that played a major role in controlling New York City and New York State politics and helping immigrants, most notably the Irish, rise up in American politics from the 1790s to the 1960s. It typically controlled Democratic Party nominations and political patronage in Manhattan from the mayoral victory of Fernando Wood in 1854 and used its patronage resources to build a loyal, well - rewarded core of district and precinct leaders; after 1850 the great majority were Irish Catholics. \\
    \dnodesm{3} & \textbf{Political party.} At the same time, the political party reached its modern form, with a membership disciplined through the use of a party whip and the implementation of efficient structures of control. The Home Rule League Party, campaigning for Home Rule for Ireland in the British Parliament was fundamentally changed by the great Irish political leader Charles Stewart Parnell in the 1880s. In 1882, he changed his party's name to the Irish Parliamentary Party and created a well-organized grass roots structure, introducing membership to replace ``ad hoc'' informal groupings. He created a new selection procedure to ensure the professional selection of party candidates committed to taking their seats, and in 1884 he imposed a firm `party pledge' which obliged MPs to vote as a bloc in parliament on all occasions. The creation of a strict party whip and a formal party structure was unique at the time. His party's efficient structure and control contrasted with the loose rules and flexible informality found in the main British parties; -- they soon came to model themselves on the Parnellite model. \\
    \bottomrule
  \end{tabularx}

  \vspace{6pt}
  \setlength{\tabcolsep}{6pt}
  \begin{tabular}{@{}ll@{}}
    \toprule
    State & Paragraphs shown, in order \\
    \midrule
    Question only    & --- \\
    Distractors only & \dnodesm{1}\,\dnodesm{2}\,\dnodesm{3} \\
    \(\{e_1\}\)       & \dnodesm{1}\,\enodesm{1}{hoplight}\,\dnodesm{3} \\
    \(\{e_2\}\)       & \enodesm{2}{hoplight}\,\dnodesm{1}\,\dnodesm{3} \\
    \(\{e_1,e_2\}\) & \enodesm{2}{hoplight}\,\dnodesm{1}\,\enodesm{1}{hoplight} \\
    \(\{e_1,e_3\}\)   & \dnodesm{1}\,\enodesm{1}{hoplight}\,\enodesm{3}{finallight} \\
    \(\{e_2,e_3\}\)   & \enodesm{2}{hoplight}\,\dnodesm{1}\,\enodesm{3}{finallight} \\
    \(\{e_1,e_2,e_3\}\) & \enodesm{2}{hoplight}\,\enodesm{1}{hoplight}\,\enodesm{3}{finallight} \\
    \bottomrule
  \end{tabular}
  \caption{Full input and all eight evidence states of the 3hop2 MuSiQue question of Table~\ref{tab:case_study}. Supporting paragraphs \(e_1\)--\(e_3\) and distractors \(d_1\)--\(d_3\) are shown verbatim. The six paragraphs of each question are shuffled once into a fixed order, here \(e_2, d_1, d_2, e_1, e_3, d_3\), and every state shows its paragraphs in that order, so a paragraph's position does not reveal the state. Every state except the question-only one shows three paragraphs, padded with distractors.}
  \label{tab:full_musique}
\end{table*}

\clearpage
\section{Ethical Considerations}
\label{app:ethics}

Our clinical experiments use MIMIC-IV-Ext Clinical Decision Making~\citep{hager2024evaluation}, a de-identified dataset accessed through PhysioNet under its data use agreement, and we do not release its data. \ours{} is a research method and is not intended for clinical use. We used Claude and ChatGPT as AI assistants in preparing this manuscript, in accordance with the ACL policy on AI writing assistance.